%% file: main.tex
\documentclass[10pt,twocolumn,letterpaper]{article}

\usepackage[pagenumbers]{cvpr} %

\input{preamble}
\definecolor{cvprblue}{rgb}{0.21,0.49,0.74}
\usepackage[pagebackref,breaklinks,colorlinks,allcolors=cvprblue]{hyperref}
\usepackage{multirow}
\usepackage{makecell}
\usepackage{tabularx}
\usepackage{booktabs}
\usepackage[table]{xcolor}
\def\paperID{29740} %
\def\confName{CVPR}
\def\confYear{2026}

\title{FishDetector-R1: Unified MLLM-Based Framework with Reinforcement Fine-Tuning for Weakly Supervised Fish Detection, Segmentation, and Counting}

\author{Yi Liu\textsuperscript{*}, Jingyu Song\textsuperscript{*}, Vedanth Kallakuri, Katherine A. Skinner\\
University of Michigan, Ann Arbor, MI USA\\
}

\begin{document}
\maketitle
\input{sec/0_abstract}
\let\thefootnote\relax\footnotetext{\textsuperscript{*}Equal contribution. Corresponding at: \texttt{jingyuso@umich.edu}}
\let\thefootnote\relax\footnotetext{This work was supported by a Propelling Original Data Science (PODS) Grant from the Michigan Institute for Data and AI in Society (MIDAS) at the University of Michigan.}

\input{sec/1_intro}

\input{sec/2_related_work}
\input{sec/3_methdology}

\input{sec/4_experiment}

\input{sec/5_conclusion}
{
    \small
    \bibliographystyle{ieeenat_fullname}
    \bibliography{main}
}

\input{sec/X_suppl}

\end{document}

%% file: sec/0_abstract.tex
\begin{abstract}
Analyzing underwater fish imagery is critical for ecological monitoring but remains difficult due to visual degradation and costly annotations. We introduce \textbf{FishDetector-R1}, a unified MLLM-based framework for fish detection, segmentation, and counting under \textbf{weak supervision}. On the DeepFish dataset, our framework achieves substantial gains over baselines, improving AP by 20\% and mIoU by 10\%, while reducing MAE by 30\% and GAME by 35\%. These improvements stem from two key components: a novel \textbf{detect-to-count} prompt that enforces spatially consistent detections and counts, and Reinforcement Learning from Verifiable Reward (\textbf{RLVR}) with a complementary scalable paradigm leveraging sparse point labels. Ablation studies further validate the effectiveness of this reward design. Moreover, the improvement generalizes well to other underwater datasets, confirming strong cross-domain robustness. Overall, FishDetector-R1 provides a reliable and scalable solution for accurate marine visual understanding via weak supervision. The project page for FishDetector-R1 is \url{https://umfieldrobotics.github.io/FishDetector-R1}.

\end{abstract}

%% file: sec/1_intro.tex
\section{Introduction}
\label{sec:intro}

Recent advances in underwater perception systems have underscored the growing need for robust visual understanding of marine environments, where visual data play a central role in ecological monitoring, fisheries management, and underwater exploration~\cite{stankus2021state,song2024turtlmap,song2025oceansim}. Effective analysis requires not only detecting fish but also performing instance-level segmentation and estimating their counts, which supports downstream tasks such as species identification, behavioral analysis, and habitat mapping. However, these tasks remain particularly challenging in underwater imagery, where low visibility, color distortion, and light scattering severely degrade the performance of conventional vision models.

\begin{figure}[htpb]
  \centering
 \includegraphics[width=1\linewidth]{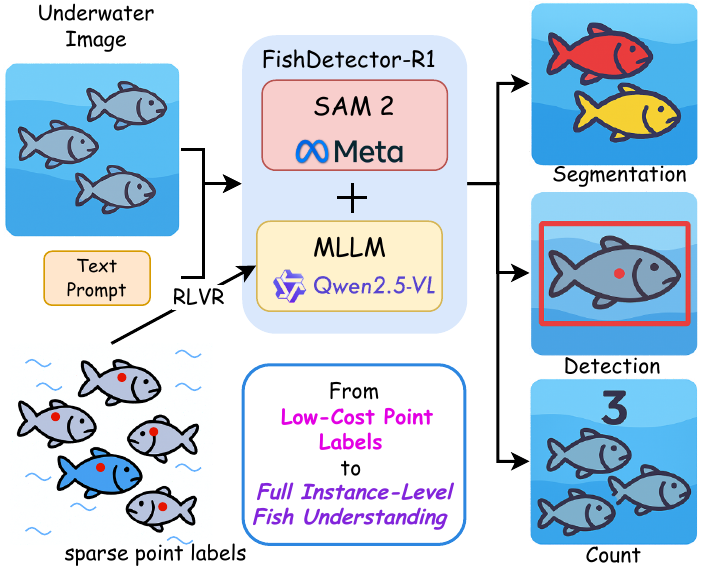}

   \caption{Our proposed FishDetector-R1 aims to achieve AI-enabled fish image analysis with the guidance of sparse point labels and text prompts.}
   \label{fig:intro_fig}
   \vspace{-2mm}
\end{figure}

Over the past decade, a range of learning-based approaches have been proposed to address these challenges. Fully supervised instance segmentation methods achieve strong performance but rely heavily on large-scale, densely annotated datasets~\cite{zhang2022dpanet, qin2016deepfish, al2022yolo, cai2018cascade}. The high cost and labor intensity of generating pixel-wise annotations make such approaches difficult to scale in underwater environments~\cite{lin2014microsoft, saleh2020realistic, lian2023watermask}. As a more annotation-efficient alternative, point-level weak supervision offers significant advantages in terms of speed and scalability~\cite{bearman2016s, laradji2018blobs}. However, existing weakly supervised methods based on point annotations often suffer from a substantial performance gap relative to fully supervised models because sparse points provide limited pixel-level guidance~\cite{laradji2020affinity, shao2022deep}.
This leaves a key question: how can we close the weak-to-dense performance gap in challenging underwater settings while still relying only on sparse, scalable point-level labels?

We address this gap with two complementary ingredients. First, we find that foundation models are well positioned to fill this gap due to their transferable visual understanding capabilities from large-scale pretraining. Multimodal large language models (MLLMs) like GPT-4 series~\cite{hurst2024gpt, achiam2023gpt, openai_gpt41_2025}, Qwen2.5-VL series~\cite{bai2025qwen2} and Gemini series~\cite{team2023gemini, comanici2025gemini} combine rich semantic knowledge with strong reasoning capability, while segmentation foundation models such as SAM~\cite{kirillov2023segment, ravi2024sam} can complement them by providing robust semantic priors for accurate mask generation from sparse prompts, enabling effective deployment in visually challenging domains.

Second, we develop an effective framework that tailors the MLLM to the specific challenges of underwater fish visual analysis. We propose a novel joint detect–to-count task formulation that turns sparse point labels into consistent, verifiable reward signals to enforce spatial alignment between the predicted detection and counting number. Building on recent successes of Reinforcement Learning from Verifiable Rewards (RLVR) for adapting foundation models~\cite{seg-zero, visionreasoner, seg-r1, visual-rft, yu2025perception}, we fine-tune the MLLM under this detect–to-count objective, yielding mutually reinforcing gains in detection and counting while supplying precise spatial priors that guide segmentation mask generation effectively. To the best of our knowledge, we are the first to integrate an MLLM with a segmentation foundation model to tackle scalable marine fish visual analysis—covering detection, instance segmentation, and counting—using only weak point-level supervision.

Together, these two ingredients constitute \textbf{FishDetector-R1} (\cref{fig:intro_fig}), a unified framework for detection, segmentation, and counting from weak point-level supervision. FishDetector-R1 moves beyond prior approaches that treat these tasks in isolation and yields concurrent improvements across all three tasks. To summarize, our contributions are as follows:

\begin{enumerate}  
   \item We propose FishDetector-R1, the first unified framework to integrate an MLLM with a segmentation foundation model for comprehensive marine fish analysis (detection, segmentation, and counting) using only weak, point-level supervision.
   \item We design a novel joint detect–to-count learning paradigm to adapt foundation models to the challenging underwater domain in a complementary manner. By formulating sparse point labels as verifiable rewards within an RLVR framework, our method enforces spatial and numerical consistency, enabling the generation of high-quality masks from minimal annotation.
   \item We conduct extensive quantitative and qualitative experiments on the DeepFish dataset~\cite{qin2016deepfish} to demonstrate the effectiveness of the proposed FishDetector-R1 pipeline, achieving performance competitive with and even surpassing fully supervised methods. We further validate its strong generalization through zero-shot transfer on another underwater dataset~\cite{islam2020suim}.

\end{enumerate}

%% file: sec/2_related_work.tex
\section{Related Work}
\label{sec:related_work}

\subsection{Fish Detection in Underwater Scenes}

Fully supervised segmentation methods~\cite{zhang2022dpanet, lian2023watermask, garcia2020automatic} achieve high accuracy in underwater scenes by training on dense pixel-wise annotations. However, such labels are time-consuming and costly to obtain, especially in underwater imagery where object boundaries are often ambiguous~\cite{saleh2020realistic, lian2023watermask}. To reduce annotation cost, weakly supervised approaches~\cite{laradji2020affinity, laradji2018blobs, shao2022deep} use point-level labels that are faster to collect~\cite{bearman2016s}, but typically yield lower segmentation performance due to the lack of dense spatial supervision, creating a clear gap between fully and weakly supervised models. While prior methods improve annotation efficiency, none have closed this gap on challenging underwater segmentation tasks. In contrast, \textbf{FishDetector-R1} is the first framework to effectively leverage only point-level supervision to achieve high-quality instance segmentation, matching or surpassing fully supervised baselines on the DeepFish benchmark~\cite{qin2016deepfish}.

\subsection{Visual Foundation Models}

Visual foundation models such as SAM~\cite{kirillov2023segment} and SAM 2~\cite{ravi2024sam} provide flexible segmentation from simple prompts like points or bounding boxes and demonstrate strong generalization across diverse visual domains. Their ability to operate in a zero-shot setting makes them attractive for domains with limited labels. Recent adaptations to underwater imagery, such as AquaSAM~\cite{xu2023aquasam} and WaterSAM~\cite{hong2024watersam}, attempt to specialize these models by either freezing encoders or introducing lightweight adapters to improve segmentation under challenging visual conditions like turbidity and color distortion. While effective, their reliance on dense supervision limits scalability and practicality in annotation-scarce scenarios. In contrast, our method effectively leverages sparse point-level labels to train a unified framework for joint detection, segmentation, and counting with reinforcement fine-tuning, supporting scalable deployment.

\label{sec:methodology}
\begin{figure*}[t]
  \centering
  \includegraphics[width=1.0\linewidth]{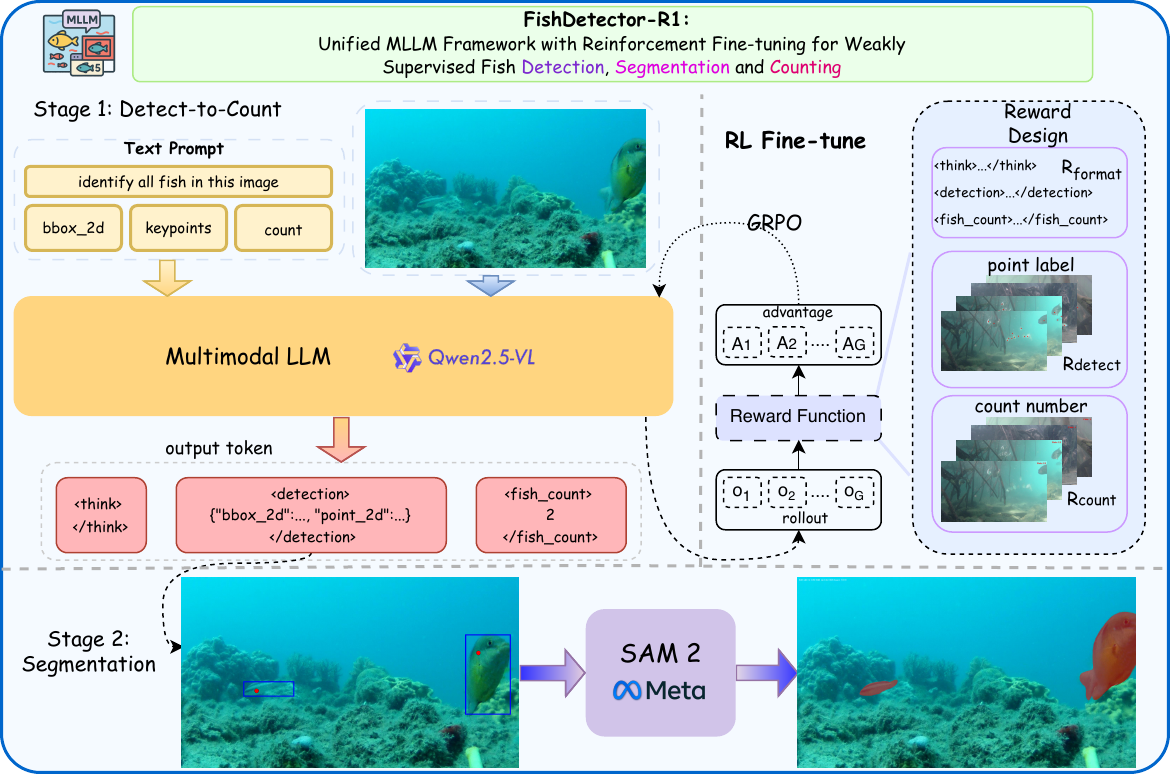}
  \caption{\textbf{Overview of the proposed FishDetector-R1 framework}. A two-stage detect-to-count pipeline integrates an MLLM with SAM 2 to jointly perform detection, segmentation, and counting. Reinforcement fine-tuning with GRPO and weak point-level supervision adapts the MLLM, ensuring consistency between detection and counting while enabling pixel-wise segmentation with only sparse labels.}
  \label{fig:pipeline}
  \vspace{-2mm}
\end{figure*}

\subsection{Multimodal Large Language Models}

MLLMs such as GPT-4.1~\cite{openai_gpt41_2025}, Llama~\cite{touvron2023llama}, and Qwen2.5-VL~\cite{bai2025qwen2} combine visual perception with natural language reasoning, enabling object grounding, spatial reasoning, and prompt-based visual interaction. These models show strong potential in general domains for zero-shot grounding and segmentation by aligning semantic priors from language instructions with visual content. However, their application to underwater imagery remains largely unexplored, even though underwater monitoring demands high-level reasoning to discern subtle visual cues. In this work, we adapt an open-source MLLM with weak point-level supervision to generate reliable bounding boxes and keypoints under noisy underwater conditions. These semantic priors then guide SAM 2 for instance-level segmentation, bridging the gap between high-level reasoning and fine-grained perception in an annotation-efficient manner.

\subsection{Reinforcement Learning for MLLMs}

Reinforcement Learning from Verifiable Reward (RLVR) has emerged as an effective strategy to improve the reasoning and alignment capabilities of both language models and multimodal models~\cite{zhang2025reinforcement}.While traditional methods like PPO~\cite{PPO} and DPO~\cite{DPO} are widely adopted, more recent approaches such as GRPO~\cite{guo2025deepseek} offer improved stability and efficiency via group preference optimization. Building on this, frameworks like Perception-R1~\cite{yu2025perception}, Seg-R1~\cite{seg-r1}, and VisionReasoner~\cite{visionreasoner} extend RL fine-tuning to multimodal perception, showing strong results in detection, segmentation, and counting. However, these works typically treat each task in isolation with separate reward functions, and focus on general-domain benchmarks, leaving domain-specific settings like underwater imagery underexplored. Our work addresses this gap by applying GRPO with point-level supervision and a unified reward design that jointly couples detection and counting. This enables all three tasks—detection, segmentation, and counting—to reinforce one another, improving adaptation under weak supervision.

%% file: sec/3_methdology.tex
\section{Methodology}
\label{sec:methdology}

\subsection{Overview}

We propose a two-stage framework, \textbf{FishDetector-R1}, that integrates an MLLM (Qwen2.5-VL~\cite{bai2025qwen2}) with a segmentation foundation model (SAM 2~\cite{ravi2024sam}) for underwater fish detection, segmentation, and counting as illustrated in \cref{fig:pipeline}. In the first stage, guided by a detect-to-count prompt, Qwen2.5-VL takes an input image, localizes each fish with a bounding box and keypoint, and then derives the total count from its detections, promoting consistency between localization and counting. In the second stage, these spatial priors are passed to SAM 2 to generate high-resolution pixel-wise instance masks. To further adapt the framework to underwater imagery, we apply RL fine-tuning to Qwen2.5-VL with weak point labels. This training step precedes SAM 2, ensuring that the MLLM learns to generate spatially consistent detections and counts, which then serve as strong priors for segmentation. Motivated by recent findings (e.g., Perception-R1~\cite{yu2025perception}), we directly apply RL fine-tuning without a preliminary supervised fine-tuning (SFT) stage. Our empirical results confirm that this strategy improves task performance while preserving annotation efficiency.

\subsection{Prompt Design}

\begin{figure}[htpb]
  \centering
 \includegraphics[width=1\linewidth]{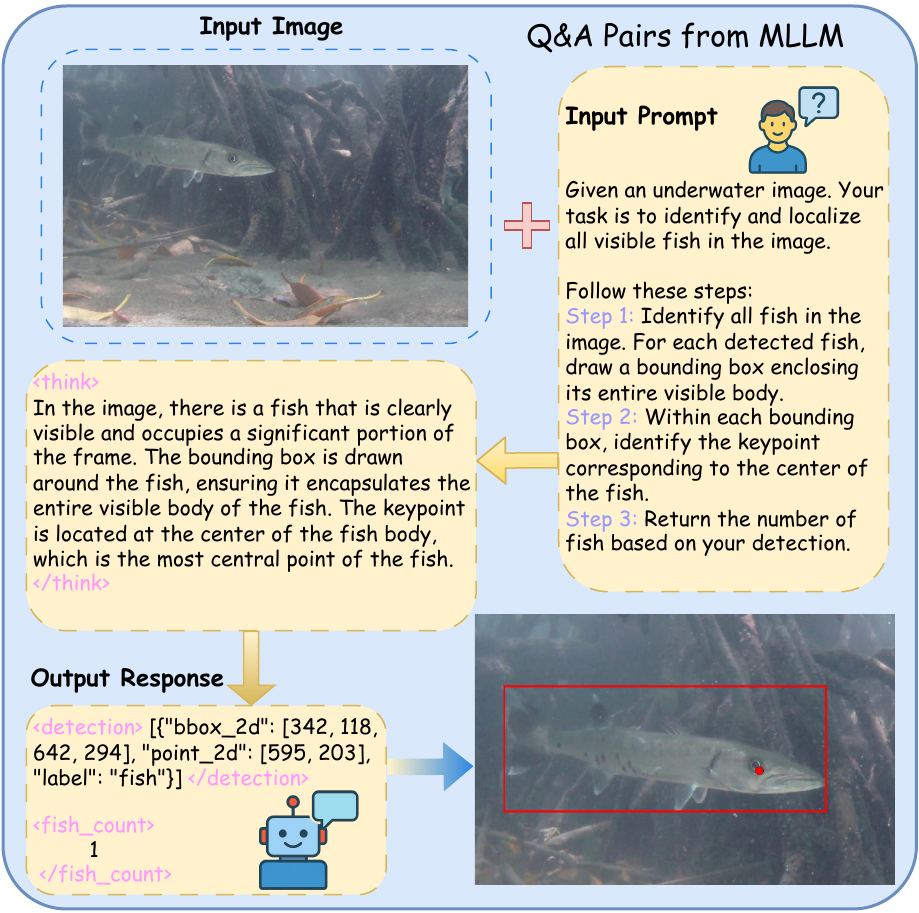}

   \caption{Example Q\&A pairs from FishDetector-R1 using our designed detect-to-count prompt.}
   \label{fig:example}
   \vspace{-5mm}
\end{figure}

To support joint detection, segmentation, and counting, we design a structured prompt tailored for \textbf{Qwen2.5-VL}, an MLLM with strong grounding and reasoning capabilities. As shown in \cref{fig:example}, given an underwater RGB image, the model is prompted to first localize each fish instance with the total count directly derived from detections, following a detect-to-count strategy. This formulation encourages the model to understand that reliable counting depends on accurate localization, i.e., it must “know where the fish are” before reporting how many fish there are (example in \cref{fig:prompt_qualitative}). The resulting detection outputs -- bounding boxes and keypoints -- are also passed as spatial priors to \textbf{SAM 2}, enabling high-quality instance segmentation. In this way, the prompt design unifies all three tasks within a single pipeline.  

We adopt a structured output format composed of three distinct components:
\texttt{\textless think\textgreater},  
\texttt{\textless detection\textgreater}, and  
\texttt{\textless fish\_count\textgreater}.  

\begin{itemize}
    \item The \texttt{\textless think\textgreater} field records the model’s internal reasoning and visual understanding process.  
    \item The \texttt{\textless detection\textgreater} field contains structured outputs for each fish instance, including a bounding box and a central keypoint, which both support counting and serve as effective prompts for SAM 2.  
    \item The \texttt{\textless fish\_count\textgreater} field provides the total number of fish, derived from the detections to ensure consistency between localization and counting.  
\end{itemize}

This design enforces a detect-to-count reasoning process, provides explicit spatial cues to guide segmentation, and ensures response completeness. Furthermore, during RL fine-tuning, the predicted count is compared against weak point-level annotations to construct reward signals, aligning detection and counting objectives without requiring dense labels.

\subsection{Group Relative Policy Optimization}

Following recent RL fine-tuning work on MLLMs~\cite{seg-zero, seg-r1, visionreasoner}, we adopt Group Relative Policy Optimization (GRPO)~\cite{guo2025deepseek} as our post-training strategy. GRPO is an efficient reinforcement learning framework that removes the need for a separate critic by directly comparing the relative quality of responses within a group. Given a task input \(t\), the current policy \(\pi_{\theta_{\text{old}}}\) generates a set of \(G\) candidate responses \(\{o_1, o_2, \ldots, o_G\}\) with corresponding rewards \(\{r_1, r_2, \ldots, r_G\}\). These rewards are normalized within the group to compute relative advantages, which are then used to update the policy. This group-wise formulation provides more stable optimization while reducing training costs compared to traditional actor–critic methods.

The GRPO objective function is defined as:

\begin{align}
\mathcal{J}_{\text{GRPO}}(\theta) = 
& \; \mathbb{E}_{o_i \sim \pi_{\theta_{\text{old}}}} \Bigg[ 
\frac{1}{G} \sum_{i=1}^{G} 
\min \Bigg(
\frac{\pi_{\theta}(o_i|t)}{\pi_{\theta_{\text{old}}}(o_i|t)} \hat{A}_i, \nonumber \\
& \quad \text{clip} \left( 
\frac{\pi_{\theta}(o_i|t)}{\pi_{\theta_{\text{old}}}(o_i|t)}, 1 - \epsilon, 1 + \epsilon 
\right) \hat{A}_i 
\Bigg) \Bigg] \nonumber \\
& - \beta D_{\text{KL}}(\pi_{\theta} \| \pi_{\text{ref}})
\label{eq:grpo}
\end{align}

\noindent where $\epsilon$ is the clipping threshold, $\beta$ is the coefficient of the KL penalty, and $\hat{A}_i$ denotes the normalized advantage for response $o_i$, computed as:

\begin{equation}
\hat{A}_i = \frac{r_i - \text{mean}(\{r_1, \ldots, r_G\})}
{\text{std}(\{r_1, \ldots, r_G\})}
\label{eq:advantage}
\end{equation}

By leveraging group-wise comparisons and reward normalization, GRPO enables stable and sample-efficient policy optimization purely based on relative preferences.
\subsection{Reward Design}

The overall reward function used for RL fine-tuning consists of four components: (1) a format reward, (2) an accuracy reward, (3) a count reward, and (4) a non-repetition reward. Each component is designed to encourage the model to produce syntactically valid, semantically accurate, and non-redundant outputs that align with weak point-level supervision.

\noindent\textbf{Format Reward.}
The format reward \( R_{\text{format}} \) has two sub-parts: response structure formatting and detection content formatting.

\begin{itemize}
    \item The response is required to contain three structured tags: 
    \texttt{\textless think\textgreater}, 
    \texttt{\textless detection\textgreater}, and 
    \texttt{\textless fish\_count\textgreater}.
    A correct structural response yields 1.0 reward.

    \item The content within the \texttt{<detection>} tag must follow the format:
    \texttt{\{"bbox\_2d": [x1, y1, x2, y2], "point\_2d": [x, y], "label": "fish"\}}.
    If all predicted instances match this structure, the model receives up to 3.0 additional reward points.
\end{itemize}

\noindent\textbf{Detection Accuracy Reward.}
To encourage correct detection and precise localization, we design an accuracy reward \(R_{\text{detect}}\). Predicted keypoints are matched to ground-truth points using the Hungarian algorithm within a Euclidean distance threshold. A prediction is considered valid if its distance to a ground-truth point is within a predefined threshold. The reward is defined as:
\vspace{-1ex}
\begin{equation}
\text{accuracy\_reward} = \lambda_{\text{detect}} \cdot \left( \frac{N_{\text{valid}}}{N_{\text{gt}}} \right),
\label{eq:accuracy_reward}
\end{equation}
    where \(N_{\text{valid}}\) denotes the number of matched predictions and \(N_{\text{gt}}\) the total number of ground-truth fish. \(\lambda_{\text{detect}}\) specifies the maximum reward assigned to the accuracy performance, which is set to 4.0 empirically. The accuracy reward scales proportionally with the fraction of correctly matched instances.

In addition, we enforce consistency between the number of detected instances \(N_{\text{pred}}\) and the reported count \(N_{\text{count}}\) in the \texttt{<fish\_count>} tag by introducing a match reward:

\vspace{-2ex}
\begin{equation}
\text{match\_reward} = 
\begin{cases}
0, & \text{if } N_{\text{pred}} = N_{\text{count}}, \\
-1, & \text{otherwise}.
\end{cases}
\label{eq:match_reward}
\end{equation}
\vspace{-2ex}

The overall detection-related reward is then computed as:
\vspace{-1ex}
\begin{equation}
R_{\text{detect}} = \text{accuracy\_reward} + \text{match\_reward}.
\end{equation}

This formulation jointly optimizes detection and counting: accurate localization improves counting reliability, while consistent counting further encourages complete detection.

\noindent\textbf{Count Reward.}
To further enforce correct enumeration, a count reward \( R_{\text{count}} \) is assigned based on whether the predicted count matches the number of ground-truth instances:
\begin{equation}
R_{\text{count}} = 
\begin{cases}
1, & \text{if } N_{\text{count}} = N_{\text{gt}} \\
-1, & \text{otherwise}
\end{cases}
\label{eq:count_reward}
\end{equation}
where \(N_{\text{count}}\) is the number of fish reported by the model and \(N_{\text{gt}}\) the total number of ground-truth fish. 

\noindent\textbf{Non-Repetition Reward.} To mitigate repetitive responses and promote output diversity, we adopt a non-repetition reward \(R_{\text{non-repeat}}\)  inspired by Seg-Zero~\cite{seg-zero}.

\noindent\textbf{Total Reward.} The total reward used for RL optimization is defined as:
\begin{equation}
R_{\text{total}} = w_1 \cdot R_{\text{format}} + w_2 \cdot R_{\text{detect}} + w_3 \cdot R_{\text{count}} + w_4 \cdot R_{\text{non-repeat}}
\end{equation}

\noindent where \(\alpha\) and \(\beta\) control the relative weight of detection and counting rewards.  
This formulation jointly accounts for syntactic correctness, localization accuracy, count fidelity, and output diversity, thereby providing rich supervision signals at minimal annotation cost.
We examine the effect of different reward–weight configurations and find that the absolute reward scale matters little, whereas the relative balance between components is critical for final performance—consistent with GRPO’s reliance on group-wise relative advantages rather than absolute magnitudes. Based on this observation, we set $w_1$, $w_2$, $w_3$, and $w_4$ to 1, as this configuration consistently yielded the strongest performance across our initial sweeps and offered a stable trade-off between detection and counting quality.

\begin{table*}[htpb]
\caption{
Unified comparison of detection ($\text{AP}_{0.5:0.95}$, $\text{AR}_{0.5:0.95}$), segmentation (Foreground, Background, mIoU), and counting (MAE, Match Rate, GAME) performance across baseline and proposed MLLM variants on the testset of DeepFish \textit{FishSeg} and \textit{FishLoc} subset. The \textbf{best} result in each column is shown in bold, and the \underline{second best} is underlined.
}
\label{tab:MLLM_all_results}
\centering
\footnotesize
\begin{tabular}{lcccccccc}
\toprule
\textbf{Model} 
& \multicolumn{2}{c}{\textbf{Detection}} 
& \multicolumn{3}{c}{\textbf{Segmentation}} 
& \multicolumn{3}{c}{\textbf{Counting}} \\
\cmidrule(lr){2-3} \cmidrule(lr){4-6} \cmidrule(lr){7-9}
& \textbf{$\text{AP}_{0.5:0.95}$ $\uparrow$} 
& \textbf{$\text{AR}_{0.5:0.95}$ $\uparrow$} 
& \textbf{Foreground $\uparrow$} 
& \textbf{Background $\uparrow$} 
& \textbf{mIoU $\uparrow$} 
& \textbf{MAE $\downarrow$} 
& \textbf{Match Rate $\uparrow$} 
& \textbf{GAME $\downarrow$} \\
\midrule
\multicolumn{9}{c}{\textbf{Baseline}} \\
\midrule
GPT-4.1 & 5.47 & 17.43 & 43.77 & 98.69 & 71.23 & \underline{0.387} & \textbf{0.796} & 1.394 \\
Gemini-2.0-flash & 54.60 & 61.24 & 84.01 & 99.64 & 91.83 & 1.879 & 0.582 & 3.434 \\
Gemini-2.5-flash & 24.10 & 46.46 & 57.14 & 98.85 & 78.00 & 2.228 & 0.568 & 3.081 \\
Qwen2.5-VL 3B & 44.63 & 62.12 & 45.10 & 97.64 & 71.67 & 0.604 & 0.616 & 1.136 \\
Qwen2.5-VL 7B & 48.05 & 55.84 & 81.25 & 99.58 & 90.42 & 0.579 & 0.706 & 0.915 \\
LangSAM & 35.37 & \textbf{68.41} & 81.33 & 99.54 & 90.43 & 1.347 & 0.323 & 1.782 \\
\midrule
\multicolumn{9}{c}{\textbf{Ours}} \\
\midrule
FishDetector-Base 3B & 53.58 & 62.21 & 67.88 & 99.10 & 83.49 & 0.647 & 0.691 & 0.901 \\ 
FishDetector-R1 3B & \textbf{61.71} & \underline{66.64} & \underline{86.47} & \underline{99.69} & \underline{93.08} & \textbf{0.386} & 0.760 & \underline{0.613} \\
\arrayrulecolor{gray!50}\midrule
FishDetector-Base 7B & 47.52 & 58.67 & 80.86 & 99.56 & 90.21 & 0.497 & 0.715 & 0.924 \\
FishDetector-R1 7B & \underline{60.71} & 63.63 & \textbf{87.90} & \textbf{99.78} & \textbf{93.84} & 0.398 & \underline{0.765} & \textbf{0.587} \\
\arrayrulecolor{black}\bottomrule
\end{tabular}
\vspace{-4mm}
\end{table*}

%% file: sec/4_experiment.tex
\section{Experiments}
\label{sec:experiment}

\subsection{Evaluation Setting}
As discussed in \cref{sec:intro}, FishDetector-R1 is a unified framework for comprehensive marine fish analysis including detection, segmentation, and counting using weak supervision signals. We evaluate FishDetector-R1 on all these tasks following standard protocals. Specifically, for segmentation, we measure mean Intersection-over-Union (mIoU) between predicted masks and ground-truth annotations. For detection, we follow the COCO evaluation protocol~\cite{lin2014microsoft} and report Average Precision (AP) and Average Recall (AR) across multiple IoU thresholds. Here we report the value of  $\text{AP}_{0.5:0.95}$ and $\text{AR}_{0.5:0.95}$, representing the mean AP and AR computed at IoU thresholds from 0.5 to 0.95.
Furthermore, to measure counting accuracy, we employ several complementary metrics. Mean Absolute Error (MAE) quantifies counting error, while the Match Rate evaluates the consistency between the predicted and ground-truth counts. Given predicted counts \(\hat{y}_i\) and ground-truth counts \(y_i\) for \(N\) images, the MAE and Match Rate are calculated as:
\vspace{-1ex}
\begin{equation}
\text{MAE} = \frac{1}{N} \sum_{i=1}^{N} \left| \hat{y}_i - y_i \right|,
\end{equation}
\vspace{-1ex}
\begin{equation}
\text{Match Rate} = \frac{1}{N} \sum_{i=1}^{N} \mathbb{1}\!\left(\hat{y}_i = y_i\right),
\end{equation}
\vspace{-1ex}

where \(\mathbb{1}(\cdot)\) is the indicator function that equals 1 if the predicted count exactly matches the ground truth and 0 otherwise.

In addition, we report the Grid Average Mean Absolute Error (GAME)~\cite{guerrero2015extremely} to compute counting errors at different spatial scales, where each image is divided into \(4^L\) non-overlapping grids at level \(L\). The error is then computed over all sub-regions:  
\vspace{-2ex}
\begin{equation}
\text{GAME}(L) = \frac{1}{N} \sum_{i=1}^{N} \sum_{r=1}^{4^L} \left| \hat{y}_i^r - y_i^r \right|,
\end{equation}
\vspace{-2ex}
\begin{equation}
\text{GAME} = \frac{1}{4} \sum_{L=1}^{4} \text{GAME}(L),
\end{equation}
\vspace{-0.5ex}

where \(\hat{y}_i^r\) and \(y_i^r\) denote the predicted keypoints and ground-truth point labels in the \(r\)-th region of the \(i\)-th image. Lower values of GAME indicate better spatial consistency in counting predictions, reducing cases where totals are correct but fish are mislocalized.

We conduct experiments using the open-source DeepFish benchmark~\cite{qin2016deepfish}. We evaluate grounding (i.e., detection and segmentation) performance on the test split of the DeepFish \textit{FishSeg} subset. For counting, we use the test split of the DeepFish \textit{FishLoc} subset. We also test FishDetector-R1 on the SUIM dataset~\cite{islam2020suim} to demonstrate the generalization capability of FishDetector-R1 across different environments and fish species in \cref{sec:suim_generalization}.

\setlength{\tabcolsep}{8pt} 
\renewcommand{\arraystretch}{1.2} 

\subsection{Implementation Details}
We build our framework based on the 3B and 7B variants of Qwen2.5-VL~\cite{bai2025qwen2}. For FishDetector-Base, we use the official pretrained Qwen2.5-VL weight with our detect-to-count prompt design. For FishDetector-R1, we apply GRPO fine-tuning on the training split of the DeepFish localization subset~\cite{qin2016deepfish}, which contains 1,600 images with point-level fish annotations. Full parameter fine-tuning on the Qwen2.5-VL is performed on 4~$\times$~NVIDIA A100 GPUs, with a batch size of 16 per device and 8 rollouts per input. Training is conducted only on the MLLM for 4 epochs with a learning rate of $1 \times 10^{-6}$. The SAM 2 is frozen and only used for segmentation evaluation. We include more implementation details in the supplementary material.

\begin{figure*}[t]
  \centering
  \includegraphics[width=1\linewidth]{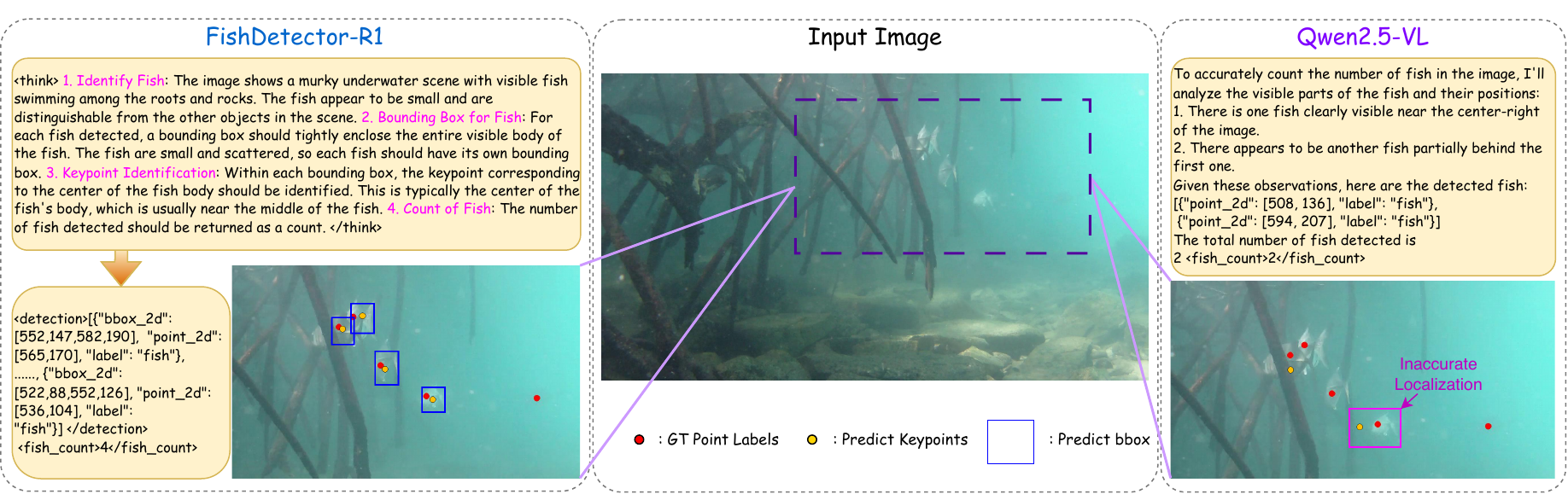}
  \caption{\textbf{Qualitative Comparison between Qwen2.5-VL and FishDetector-R1.} On a challenging scene from DeepFish \textit{FishLoc}, our detect-to-count strategy enables more accurate localization and structured outputs.}
  \label{fig:prompt_qualitative}
  \vspace{-5mm}
\end{figure*}

\subsection{Experimental Results}
\label{sec:experiment_results}

\subsubsection{Overall Comparison}
\textbf{Baselines.} We benchmark our framework against state-of-the-art MLLMs, including GPT-4.1~\cite{openai_gpt41_2025}, Gemini-2.0/2.5~\cite{team2023gemini, comanici2025gemini}, and Qwen2.5-VL~\cite{bai2025qwen2}, across detection, segmentation, and counting. Specifically, we prompt the MLLMs to generate coordinates of bounding boxes, which are then passed to SAM 2~\cite{ravi2024sam} to further generate segmentation masks. The MLLMs are also instructed to output the total number of fish, serving as their predictions for the counting task. The prompts are designed to be compact and general. We use the same prompts for these MLLMs to ensure fair comparison. Detailed prompts can be found in the supplementary material.
Additionally, we compare it against LangSAM~\cite{medeiros2025langsegmentanything}, a representative open-source model capable of open-set detection and segmentation by leveraging SAM 2~\cite{ravi2024sam} and Grounding DINO~\cite{liu2024grounding}.

\noindent\textbf{Quantitative Results.} The overall comparison is shown in \cref{tab:MLLM_all_results}. Our model achieves consistent and substantial improvements over all baselines across the three tasks. Notably, FishDetector-Base, which applies the detect-to-count prompt without any gradient-based fine-tuning, already yields clear gains over Qwen2.5-VL, confirming the effectiveness of our prompt design. Building on this, FishDetector-R1 further surpasses FishDetector-Base in all tasks, demonstrating the benefits of our proposed reward formulation and RLVR pipeline. We also observe that the 3B variant of FishDetector-Base gains more from the detect-to-count prompting than the 7B variant, suggesting that larger models benefit less from prompt-only adaptation and thus require additional alignment (e.g., RL fine-tuning) to fully realize performance gains. While LangSAM achieves the highest AR and the fourth highest mIoU, the poor detection precision and counting performance highlight the significant limitation of misdetection. The model struggles to distinguish fish-like objects in the background, undermining its reliability.

\noindent\textbf{Qualitative Comparison.}
\Cref{fig:prompt_qualitative} presents a comparison with the original Qwen2.5-VL in a challenging crowded scene, illustrating that FishDetector-R1 maintains stronger performance even under difficult conditions. We present more qualitative results in the supplementary material.

\begin{table}[t]
    \caption{Comparison of segmentation accuracy (mIoU) on DeepFish \textit{FishSeg} dataset across different supervision methods. The \textbf{best} result in each column is shown in bold, and the \underline{second best} is underlined.}
    \label{tab:miou_supervision}
    \centering
    \footnotesize
    \begin{tabular}{lcc}
        \toprule
        \textbf{Method} & \textbf{Supervision Type} & \textbf{mIoU} \\
        \midrule
        DeepFish~\cite{saleh2020realistic} & Dense Annotations & 93.0 \\
        A-LCFCN~\cite{laradji2020affinity} & Point Labels & 86.2 \\
        FishDetector-R1 3B & Point Labels & \underline{93.1} \\
        FishDetector-R1 7B & Point Labels & \textbf{93.8} \\
        \bottomrule
    \end{tabular}
    \vspace{-3mm}
\end{table}

\subsubsection{Comparison with Traditional Fully \& Weakly Supervised Methods}
To complete our evaluation, we compare FishDetector-R1 with both fully and weakly supervised baselines. For the fully supervised setting, we report the DeepFish benchmark~\cite{saleh2020realistic}, which trains a segmentation model using dense pixel-level masks with a ResNet-50 backbone~\cite{he2016deep}. For the weakly supervised setting, we include A-LCFCN~\cite{laradji2020affinity}, a state-of-the-art point-supervised method trained on the same DeepFish \textit{FishLoc} subset.

As shown in \cref{tab:miou_supervision}, there is a clear performance gap between weak and dense supervision (86.2 vs. 93.0 mIoU). Remarkably, FishDetector-R1 closes this gap entirely. With only sparse point annotations, the 3B variant matches the fully supervised baseline, while the 7B variant surpasses it. This demonstrates that our approach effectively combines weak supervision with the spatial reasoning and semantic priors inherent in foundation models. By leveraging multimodal prompting and reinforcement alignment, our framework not only exceeds traditional weakly supervised methods but also rivals dense annotation-based models, offering a scalable and annotation-efficient solution for underwater segmentation.

\begin{table}[t]
    \caption{Comparison of segmentation accuracy (mIoU) on the \textit{Fish (and Vertebrates)} (FV) category of SUIM dataset~\cite{islam2020suim}. The \textbf{best} result in each column is shown in bold, and the \underline{second best} is underlined.}
    \label{tab:miou_SUIM}
    \centering
    \footnotesize
    \begin{tabular}{lccc}
        \toprule
        \textbf{Method} & \textbf{Foreground} &\textbf{Background} & \textbf{mIoU} \\
        \midrule
        LangSAM & 66.20 & 96.51 & 81.35\\
        Qwen2.5-VL 3B & 30.74 & 91.96  & 61.35\\
        Qwen2.5-VL 7B & 30.33 & 93.05 & 61.69\\
        FishDetector-Base 3B & 37.85 & 95.46 & 66.66 \\
        FishDetector-Base 7B & 46.76 & 94.86 & 70.81 \\
        FishDetector-R1 3B & \textbf{70.18} & \underline{97.66} & \textbf{83.92}  \\
        FishDetector-R1 7B & \underline{69.67} & \textbf{97.72} & \underline{83.70} \\
        \bottomrule
    \end{tabular}
    \vspace{-2mm}
\end{table}

\begin{table*}[htpb]
\caption{
Ablation study of FishDetector-R1 (7B) with different reward configurations. Detection ($\text{AP}_{0.5:0.95}$, $\text{AR}_{0.5:0.95}$), segmentation (Foreground, Background, mIoU), and counting (MAE, Match Rate, GAME) metrics are reported. The \textbf{best} result in each column is shown in bold, and the \underline{second best} is underlined.
}
\label{tab:reward_ablation_all}
\centering
\footnotesize
\begin{tabular}{lcccccccc}
\toprule
\textbf{Reward Setting} 
& \multicolumn{2}{c}{\textbf{Detection}} 
& \multicolumn{3}{c}{\textbf{Segmentation}} 
& \multicolumn{3}{c}{\textbf{Counting}} \\
\cmidrule(lr){2-3} \cmidrule(lr){4-6} \cmidrule(lr){7-9}
& \textbf{$\text{AP}_{0.5:0.95}$} 
& \textbf{$\text{AR}_{0.5:0.95}$} 
& \textbf{Foreground } 
& \textbf{Background } 
& \textbf{mIoU} 
& \textbf{MAE} 
& \textbf{Match Rate} 
& \textbf{GAME } \\
\midrule
Base & 47.52 & 58.67 & 80.86 & 99.56 & 90.21 & 0.497 & 0.715 & 0.924 \\
+$R_{\text{count}}$ & 26.46 & 39.91 & 51.71 & 98.59 & 87.82 & \underline{0.414} & \textbf{0.770} & 1.346 \\
+$R_{\text{detect}}$ & \underline{57.10} & \underline{63.00} & \underline{87.10} & \underline{99.68} & \underline{93.41} & 0.442 & 0.757 & \underline{0.693} \\
+$R_{\text{count}}$+$R_{\text{detect}}$ & \textbf{60.71} & \textbf{63.63} & \textbf{87.94} & \textbf{99.78} & \textbf{93.84} & \textbf{0.398} & \underline{0.765} & \textbf{0.587} \\
\bottomrule
\end{tabular}
\vspace{-2mm}
\end{table*}

\subsection{Generalization Experiment}
\label{sec:suim_generalization}
In addition to experiments in \cref{sec:experiment_results}, we further evaluate FishDetector-R1 on an additional underwater fish dataset (i.e., SUIM~\cite{islam2020suim}) to assess the generalizability to new environments and new fish species. Specifically, we run zero-shot inference using FishDetector-Base and FishDetector-R1 on the SUIM dataset and report segmentation performance in \cref{tab:miou_SUIM}. As shown in \cref{tab:miou_SUIM}, FishDetector-R1 substantially outperforms both the base models and prior multimodal baselines. The performance gain is consistent across both 3B and 7B variants, highlighting the robustness of our reinforcement alignment and detect-to-count prompt design. These results indicate that the improvements observed on DeepFish generalize well to new underwater environments, confirming the scalability and adaptability of our framework beyond the training distribution.

\subsection{Ablation Study}
We further conduct ablation studies to examine the role of each proposed reward component. In this section, we report results of the 7B model, with additional ablations provided in the supplementary material.
\label{sec:ablation_study}

\noindent\textbf{Reward Design.}
As shown in \cref{tab:reward_ablation_all}, using only the count reward $R_{\text{count}}$ improves overall count accuracy (lower MAE and higher Match Rate), but detection and segmentation scores drop noticeably. This indicates that the model learns to match the total number of objects but does not reliably localize them in the image.
Using only the detection reward $R_{\text{detect}}$ leads to the opposite trend. Detection and segmentation performance improve (higher AP/AR and mIoU), and spatial counting consistency also improves (lower GAME). However, the total count is still not as accurate as in the $R_{\text{count}}$ setting.

When both rewards are used together, the model achieves the best overall results. Detection and segmentation reach their highest values, and counting is both more accurate (lowest MAE) and more spatially consistent (lowest GAME). This shows that these two rewards are complementary: $R_{\text{count}}$ provides high-level supervision on object cardinality, while $R_{\text{detect}}$ offers fine-grained guidance for precise localization. Combining them leads to stronger performance across detection, segmentation, and counting.

\begin{table}[t]
\caption{Alignment rate between detected instances and predicted fish counts from model output response. Higher is better.}
\label{tab:alignment_rate}
\centering
\footnotesize
\begin{tabular}{lcc}
\toprule
\textbf{Model} & \multicolumn{2}{c}{\textbf{Alignment Rate (\%)}$\uparrow$} \\
\cmidrule(lr){2-3}
 & \textbf{3B} & \textbf{7B} \\
\midrule
FishDetector-Base & 97.2 & 98.6 \\
FishDetector-R1 & \textbf{99.6} & \textbf{100} \\
\bottomrule
\end{tabular}
\vspace{-5mm}
\end{table}

\noindent\textbf{Internal Detect-to-Count Consistency}
We evaluate how consistently the model's predicted fish count matches the number of detected instances. As shown in \cref{tab:alignment_rate}, FishDetector-Base already produces high agreement, but small mismatches remain. After RL fine-tuning, FishDetector-R1 achieves near-perfect alignment for both 3B and 7B variants. This indicates that the reward-based training not only improves accuracy, but also strengthens the internal link between detection and counting, leading to more stable and reliable outputs for downstream ecological analysis.

%% file: sec/5_conclusion.tex
\section{Limitations and Future Work}
\label{sec:limits}

While FishDetector-R1 achieves notable gains, several limitations remain. First, the computational overhead of large MLLMs makes real-time deployment on resource-constrained edge devices challenging. Future work will explore quantization and edge-optimization to improve efficiency. Second, the framework can still hallucinate spurious detections or counts in cluttered scenes, highlighting the need for uncertainty modeling or verification mechanisms to enhance reliability.
Third, the current framework operates on single-frame inputs, limiting its ability to leverage temporal cues for tracking or resolving occlusions that are common in underwater environments. Future extensions will incorporate temporal modeling to improve robustness in continuous or video-based monitoring.
Finally, although FishDetector-R1 is class-agnostic by design, in this study we restrict supervision and evaluation to \emph{fish-only} detection and segmentation, which covers a relatively narrow range of habitats and species. Extending to multi-class and species-level settings (including non-fish marine life and anthropogenic objects) is an important step toward broader ecological impact.

\section{Conclusion}
\label{sec:conclusion}

We propose \textbf{FishDetector-R1}, a unified framework for underwater fish detection, segmentation, and counting that combines an MLLM with SAM 2. Through detect-to-count prompting and reinforcement fine-tuning with sparse point labels, our method achieves strong performance across tasks on the DeepFish dataset. Notably, FishDetector-R1 bridges the gap between traditional weak and fully supervised models, delivering high-quality pixel-wise segmentation with minimal annotation cost.  The improvement further generalizes to other underwater datasets, demonstrating robust cross-domain scalability. This enables scalable, annotation-efficient fish analysis, supporting real-world applications in ecological monitoring and marine habitat assessment.

%% file: sec/X_suppl.tex
\clearpage
\maketitlesupplementary

In this material, we provide additional results and analyses to further substantiate the effectiveness and practicality of FishDetector\text{-}R1. We first describe additional implementation details, including the RL training setup, prompt design, and open-source configuration used in our experiments. We then present a series of supplementary ablations: (i) reward configuration studies on the 3B base model, (ii) sensitivity analysis over reward weight choices, and (iii) the impact of different SAM~2 backbone sizes. Next, we include extended qualitative results on DeepFish~\cite{saleh2020realistic} for both segmentation and point-level localization, as well as zero-shot transfer visualizations on SUIM~\cite{islam2020suim}, highlighting robustness across habitats and domains. Finally, we discuss potential negative societal impacts of our framework in large-scale ecological and fisheries applications.

\section{Additional Implementation Details}
We provide additional implementation details to supplement the main paper. We implement and train FishDetector-R1 based on Easy-R1~\cite{zheng2025easyr1}, an open-source RL training framework. For all the instances of FishDetector-R1, we set the temperature value as the default value $1.0$. 
Furthermore, \cref{tab:prompt_comparison} summarizes the prompts used during evaluation for both the MLLM baseline~\cite{bai2025qwen2} and our FishDetector variants. For Qwen2.5-VL, we follow the official documentation: the grounding task uses a detection-style prompt to generate bounding boxes that are then passed to SAM~2~\cite{ravi2024sam} for segmentation, while the counting task adopts a point-based strategy that outputs both the total number of fish and their 2D keypoint locations for evaluating localization accuracy. In contrast, our \textit{detect-to-count} prompt in FishDetector-Base/R1 unifies grounding and counting within a single instruction, aiming to enforce consistency between localized detections and global count predictions.
For other MLLM baselines (e.g., GPT-4.1~\cite{openai_gpt41_2025}, Gemini-2.0/2.5~\cite{team2023gemini, comanici2025gemini}), we found that their outputs were somewhat sensitive to the exact same prompt. To ensure a fair comparison and allow each baseline to perform optimally, we applied minor prompt adjustments tailored to each baseline. All baseline prompts will be released with our open-source package. In addition, we will release FishDetector-R1 as fully open-source to benefit the broader community.

\begin{table*}[htpb]
\centering
\footnotesize
\caption{Prompts used for MLLM baseline~\cite{bai2025qwen2} and our FishDetector. 
Baseline separates grounding and counting, while our detect-to-count prompt unifies them for consistency.}
\begin{tabularx}{\textwidth}{@{}l|l|X|X@{}}
\toprule
\textbf{Method} & \textbf{Task} & \textbf{Prompt} & \textbf{Example} \\
\midrule
\multirow{2}{*}{Qwen2.5-VL} 
 & Grounding & Detect all fish in the image and return their locations in the form of bounding box coordinates.  
 & \texttt{[\{"bbox\_2d": [x1, y1, x2, y2], "label": "fish"\}]} \\
\cmidrule(lr){2-4}
 & Counting  & Count the number of fish in the image, including those that are only partially visible. 
 First detect their keypoints, then output the total count in \texttt{<fish\_count></fish\_count>}. 
 & \texttt{[\{"point\_2d": [x, y], "label": "fish"\}], <fish\_count>1</fish\_count>}  \\
\midrule
\multirow{2}{*}{\makecell[l]{FishDetector-\\Base / R1}} 
 & \multirow{2}{*}{\makecell[l]{Grounding \\ and \\ Counting}} 
 & Given an underwater image, identify all fish. Step 1: Draw a bounding box tightly around each fish. 
 Step 2: Mark the keypoint at the fish center. Step 3: Return the number of fish consistent with detections. 
 & \texttt{<think>...</think> 
 <detection>[\{"bbox\_2d": [x1,y1,x2,y2], "point\_2d": [x,y], "label": "fish"\}]</detection> 
 <fish\_count>1</fish\_count>} \\
\bottomrule
\end{tabularx}

\label{tab:prompt_comparison}
\end{table*}

\begin{table*}[htpb]
\caption{
Ablation study of FishDetector-R1 (3B) with different reward configurations. Detection ($\text{AP}_{0.5:0.95}$, $\text{AR}_{0.5:0.95}$), segmentation (Foreground, Background, mIoU), and counting (MAE, Match Rate, GAME) metrics are reported. The \textbf{best} result in each column is shown in bold, and the \underline{second best} is underlined.
}
\label{tab:reward_ablation_3B}
\centering
\footnotesize
\begin{tabular}{lcccccccc}
\toprule
\textbf{Reward Setting} 
& \multicolumn{2}{c}{\textbf{Detection}} 
& \multicolumn{3}{c}{\textbf{Segmentation}} 
& \multicolumn{3}{c}{\textbf{Counting}} \\
\cmidrule(lr){2-3} \cmidrule(lr){4-6} \cmidrule(lr){7-9}
& \textbf{$\text{AP}_{0.5:0.95}$ $\uparrow$} 
& \textbf{$\text{AR}_{0.5:0.95}$ $\uparrow$} 
& \textbf{Foreground $\uparrow$} 
& \textbf{Background $\uparrow$} 
& \textbf{mIoU $\uparrow$} 
& \textbf{MAE $\downarrow$} 
& \textbf{Match Rate $\uparrow$} 
& \textbf{GAME $\downarrow$} \\
\midrule
Base & 53.6 & 62.2 & 67.9 & 99.1 & 83.5 & 0.647 & 0.691 & 0.901 \\
+$R_{\text{count}}$ & 5.0 & 18.3 & 24.4 & 97.6 & 61.0 & \underline{0.415} & \textbf{0.768} & 1.646 \\
+$R_{\text{detect}}$ & \underline{59.8} & \underline{65.2} & \textbf{86.7} & \underline{99.7} & \textbf{93.2} & 0.442 & 0.757 & \underline{0.693} \\
+$R_{\text{count}}$+$R_{\text{detect}}$ & \textbf{61.7} & \textbf{66.6} & \underline{86.5} & \textbf{99.7} & \underline{93.1} & \textbf{0.398} & \underline{0.765} & \textbf{0.587} \\
\bottomrule
\end{tabular}
\vspace{-2mm}
\end{table*}

\section{Supplementary Experiments}
\subsection{Ablation on Reward Configuration with 3B Base Model}
To further validate the robustness of our approach, we report ablation experiments on the 3B model, demonstrating that FishDetector-R1 yields consistent improvements independent of the underlying model variant. As shown in \cref{tab:reward_ablation_3B}, reward design plays a decisive role in shaping overall performance. While adding the counting reward alone (\(R_{\text{count}}\)) improves overall count accuracy with less MAE and higher Match Rate, it substantially degrades both detection and segmentation accuracy and leads to worst localization precision, indicating that this signal is insufficient on its own to guide precise grounding. In contrast, introducing the detection reward (\(R_{\text{detect}}\)) produces substantial gains across all metrics, increasing AP from 53.6 to 59.8 and mIoU from 83.5 to 93.2, underscoring the importance of explicit localization feedback. Combining \(R_{\text{detect}}\) and \(R_{\text{count}}\) yields the strongest detection performance and competitive segmentation quality, and achieves more accurate (lowest MAE) and spatial consistent counting results(lowest GAME),  suggesting that the two rewards offer complementary benefits—where \(R_{\text{detect}}\) provides the primary grounding signal and \(R_{\text{count}}\) contributes additional structural guidance. These findings confirm the effectiveness of our reward formulation and the stability of FishDetector-R1 across configurations.

\subsection{Ablation on Reward Weight Configuration}

To understand how different reward components contribute to RL fine-tuning, we analyze the effect of varying the reward weights in the total objective. As described in the main paper, the reward function integrates four complementary signals—format correctness, detection accuracy, counting consistency, and non-repetition—to encourage the model to produce structured, accurate, and non-redundant outputs under weak point-level supervision. The overall formulation is:

\begin{equation}
R_{\text{total}} = w_1 \cdot R_{\text{format}}
                + w_2 \cdot R_{\text{detect}}
                + w_3 \cdot R_{\text{count}}
                + w_4 \cdot R_{\text{non-repeat}}.
\end{equation}

In this ablation, we focus on the weights directly related to task performance, \(w_2\) and \(w_3\), while keeping \(w_1\) and \(w_4\) fixed because they regulate structural formatting and repetition behavior that are irrelevant to our detection or counting tasks. Full definitions and implementation details for each reward component are provided in the main paper; here, we present an extended analysis of how the choice of \(w_2\) and \(w_3\) influences detection and segmentation performance based on 3B model. As shown in \cref{tab:weight_config_3B}, increasing the relative weight\(w_2\) enhance segmentation quality but leads to a decline in counting accuracy and localization robustness. In contrast, emphasizing \(w_3\) yields gains on the overall counting accuracy but this comes at the cost of compromising detection precision, localization precision, and segmentation performance. The proposed configuration represents an optimal compromise, offering the strongest localization capability and highly competitive detection performance, with minimal performance loss across other metrics.
\begin{table*}[htpb]
\caption{
Ablation study of FishDetector-R1 (3B) with different different weight configuration ($w_2$ : $w_3$). Detection ($\text{AP}_{0.5:0.95}$, $\text{AR}_{0.5:0.95}$), segmentation (Foreground, Background, mIoU), and counting (MAE, Match Rate, GAME) metrics are reported.}
\label{tab:weight_config_3B}
\centering
\footnotesize
\begin{tabular}{lcccccccc}
\toprule
\textbf{Weight Configuration} 
& \multicolumn{2}{c}{\textbf{Detection}} 
& \multicolumn{3}{c}{\textbf{Segmentation}} 
& \multicolumn{3}{c}{\textbf{Counting}} \\
\cmidrule(lr){2-3} \cmidrule(lr){4-6} \cmidrule(lr){7-9}
\textbf{$w_2 : w_3$}
& \textbf{$\text{AP}_{0.5:0.95}$ $\uparrow$} 
& \textbf{$\text{AR}_{0.5:0.95}$ $\uparrow$} 
& \textbf{Foreground $\uparrow$} 
& \textbf{Background $\uparrow$} 
& \textbf{mIoU $\uparrow$} 
& \textbf{MAE $\downarrow$} 
& \textbf{Match Rate $\uparrow$} 
& \textbf{GAME $\downarrow$} \\
\midrule
\multicolumn{9}{c}{FishDetector-R1 3B} \\
\arrayrulecolor{gray!50}\midrule
4 : 1 & 62.8 & 65.3  & 87.0 & 99.7 & 93.4 & 0.534 & 0.747 &  0.717\\
2 : 1 & 59.7 & 66.1 & 87.4 & 99.7 & 93.6 & 0.395 & 0.774 & 0.626\\
1 : 1 (Ours) & 61.7 & 66.6 & 86.5 & 99.7 & 93.1 & 0.398 & 0.765 & 0.587 \\
1 : 2 & 53.6 & 62.2 & 86.4 & 99.7 & 93.0 & 0.375 & 0.774 & 0.650 \\
1 : 4 & 56.9 & 63.9 & 85.5 & 99.6 & 92.6 & 0.373 & 0.776 & 0.631 \\
\arrayrulecolor{black}\bottomrule
\end{tabular}
\vspace{-2mm}
\end{table*}

\subsection{Effects on Different Sizes of SAM 2 Model}
To understand how the choice of segmentation backbone influences overall performance, we evaluate FishDetector-R1 using four different sizes of the SAM~2 model, as summarized in \cref{tab:sam2size}. The results show that SAM~2-Large is consistently the strongest choice across both the 3B and 7B variants of FishDetector-R1, achieving the highest foreground accuracy, background accuracy, and overall mIoU in every setting. This validates our decision to use SAM~2-Large as the default mask generation model in the main experiments. At the same time, the performance degradation when moving to smaller SAM~2 models is modest—particularly for the 7B variant—indicating that lighter backbones still deliver satisfactory segmentation quality. This suggests that FishDetector-R1 can flexibly support resource-constrained or edge deployments while retaining strong performance.
\begin{table}[h]
\centering
\footnotesize
\caption{Comparison of segmentation quality with different size of SAM 2 model.}
\begin{tabularx}{\linewidth}{lXXX}
\toprule
\textbf{SAM 2 Model Size} & \textbf{Foreground} & \textbf{Background} & \textbf{mIoU}\\
\midrule
\multicolumn{4}{c}{FishDetector-R1 3B} \\
\arrayrulecolor{gray!50}\midrule
SAM 2-Large & 86.47 & 99.69 & 93.08\\
SAM 2-Base & 84.39 & 99.64 &  92.02\\
SAM 2-Small & 84.52 & 99.65 & 92.09 \\
SAM 2-Tiny & 83.03 & 99.61 & 91.32 \\
\midrule
\multicolumn{4}{c}{FishDetector-R1 7B} \\
\midrule
SAM 2-Large & 87.90 & 99.78 & 93.84\\
SAM 2-Base & 86.95 & 99.71 &  93.33\\
SAM 2-Small & 86.26 & 99.70 & 92.98 \\
SAM 2-Tiny & 86.39 & 99.70 & 93.05 \\
\arrayrulecolor{black}\bottomrule
\label{tab:sam2size}
\end{tabularx}
\end{table}

\subsection{Qualitative Results}
\subsubsection{Robustness in Diverse Underwater Habitats}
\begin{figure*}[t]
  \centering
  \includegraphics[width=1\linewidth]{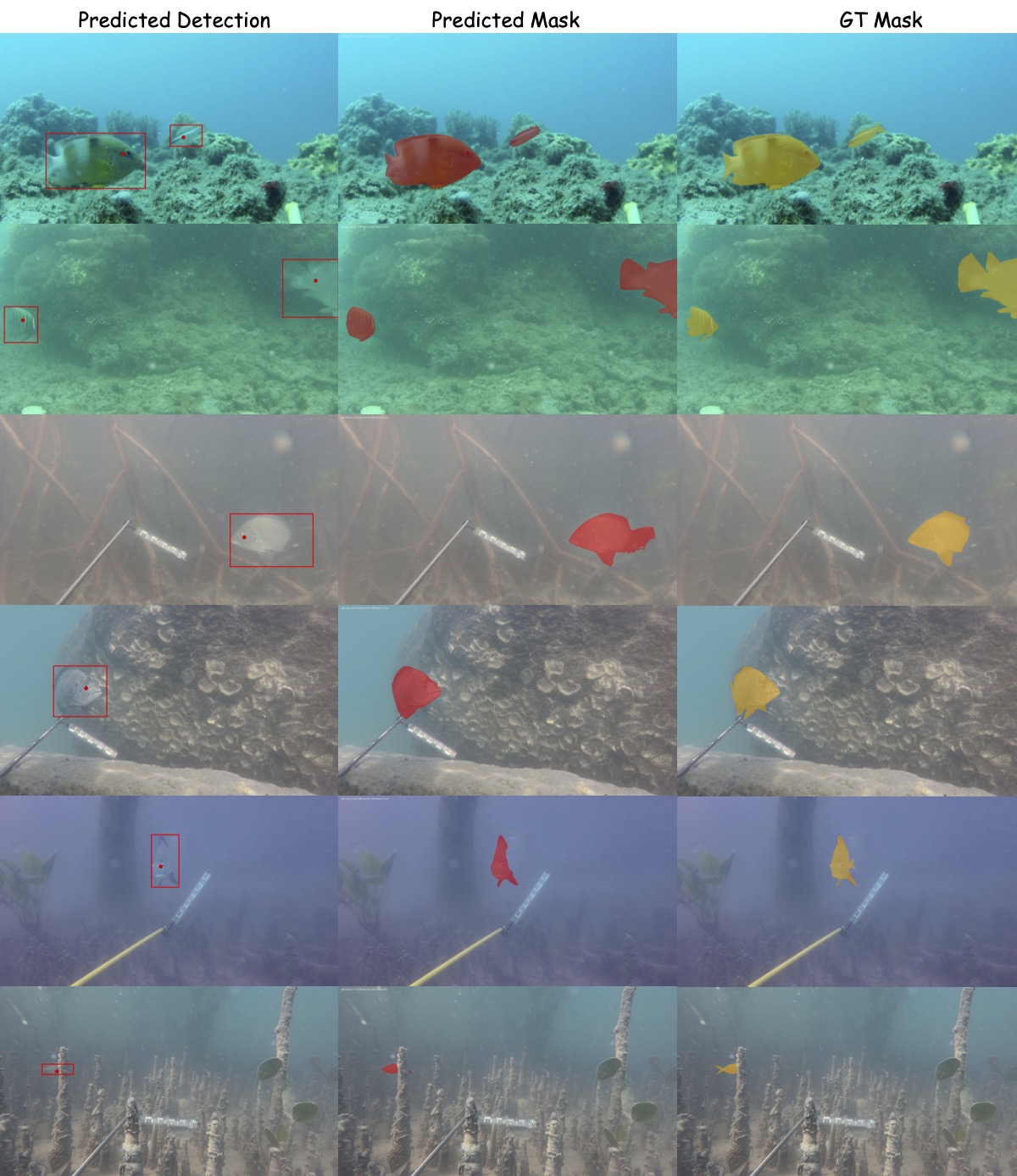}
  \caption{Detection and segmentation results of FishDetector-R1 across diverse underwater habitats. 
  The model demonstrates robustness to variations in background complexity, lighting conditions, and water color, highlighting its applicability to real-world marine environments.}
  \label{fig:seg_quality}
\end{figure*}

Figure \ref{fig:seg_quality} illustrates qualitative detection and segmentation results of our FishDetector-R1 framework across diverse underwater habitats from the DeepFish dataset~\cite{saleh2020realistic}. The selected examples span a variety of challenging conditions, including complex coral reefs, turbid shallow waters, and scenes with varying illumination and water color. Despite these challenges, the model successfully detects fish that are visually similar to the background, partially occluded, or swimming close to fish-like distractors. Moreover, the predicted segmentation masks are fine-grained and closely aligned with ground-truth labels, demonstrating the ability of FishDetector-R1 to capture accurate object boundaries under degraded visual conditions.  

These results highlight not only the robustness of our framework in handling environmental variability but also its practical significance: reliable detection and segmentation of fish across habitats is essential for downstream ecological applications such as species identification, population estimation, and habitat health assessment. By reducing the reliance on dense annotations while maintaining strong performance, FishDetector-R1 offers a scalable tool for large-scale ecological monitoring and sustainable fisheries management in real-world marine environments.

\subsubsection{Improving Count Accuracy with Spatially Aligned Keypoints}
\begin{figure*}[t]
  \centering
  \includegraphics[width=1\linewidth]{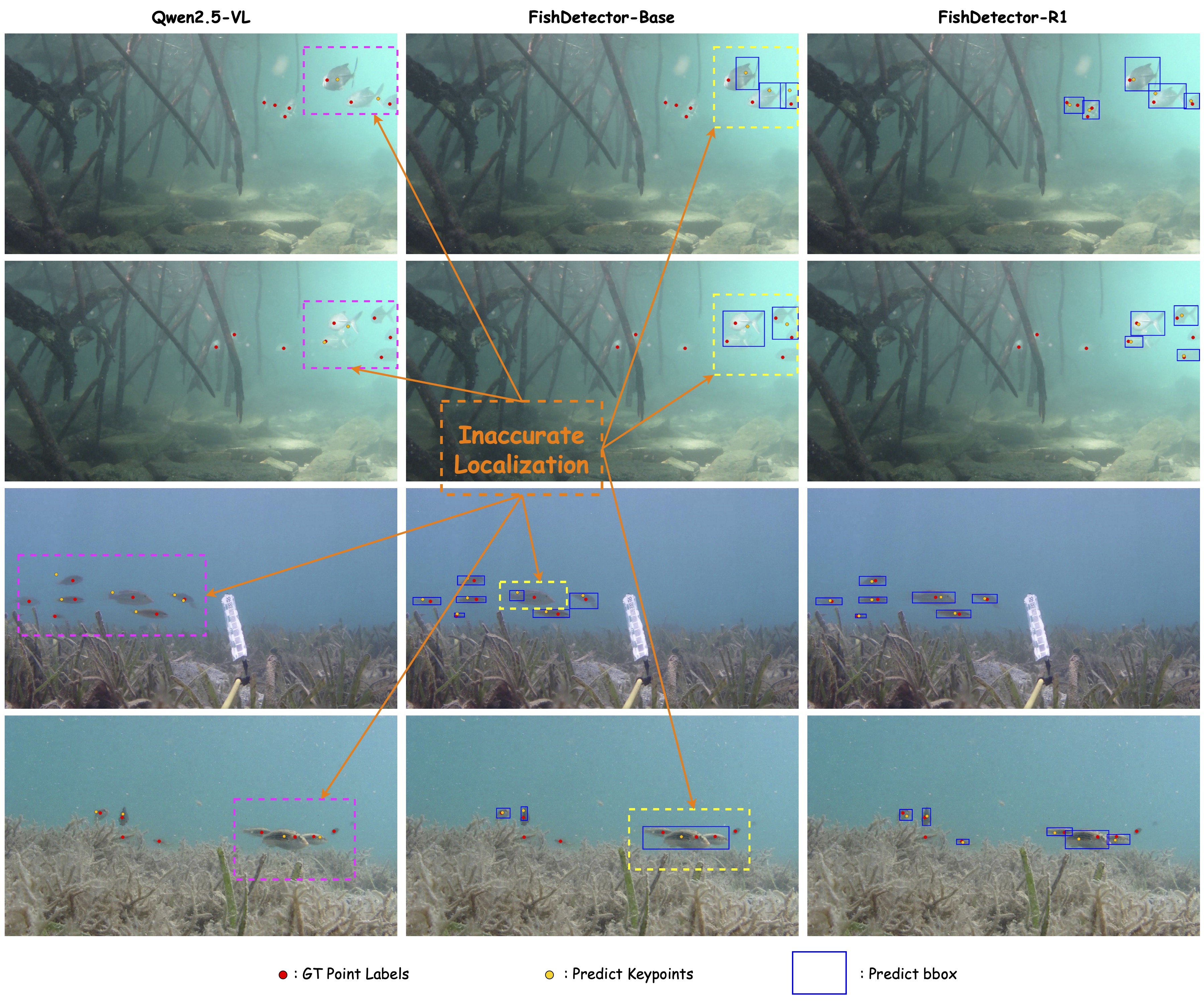}
  \caption{Qualitative detection results on the DeepFish \textit{FishLoc} test set. In crowded scenes with many small fish, FishDetector-R1 delivers more accurate counts and tighter localizations. Best viewed in color and zoomed in.}
  \label{fig:count_comparison_2}
\end{figure*}

\begin{figure*}[t]
  \centering
  \includegraphics[width=1\linewidth]{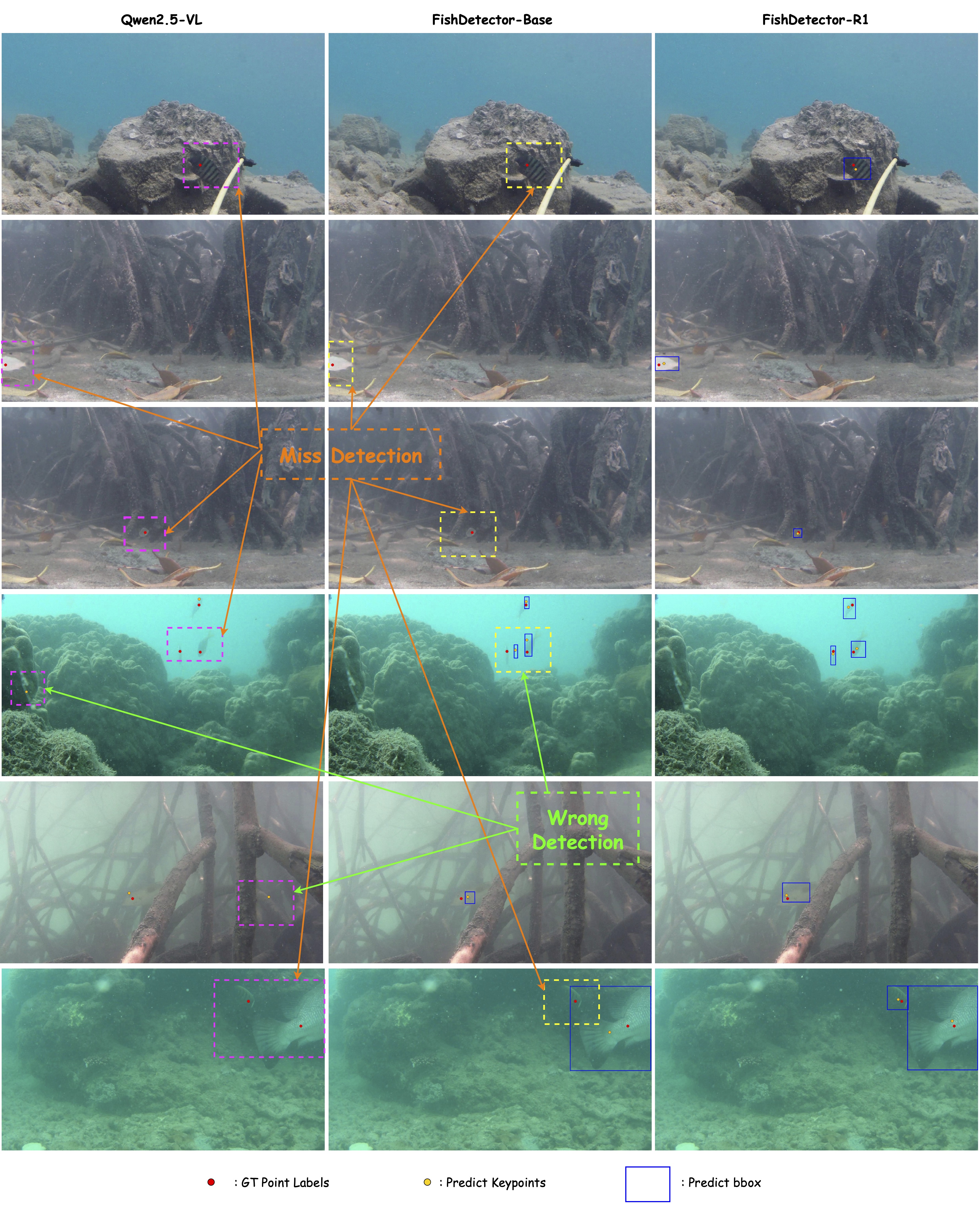}
\caption{Qualitative detection results on the DeepFish \textit{FishLoc} test set. FishDetector-R1 yields the most accurate and robust detections, especially under strong external disturbances. Best viewed in color and zoomed in.}
  \label{fig:count_comparison_1}
\end{figure*}

In \cref{fig:count_comparison_2,fig:count_comparison_1}, we present challenging scenes from the DeepFish \textit{FishLoc} test set~\cite{saleh2020realistic} under point-level supervision: (i) crowded schools of small, closely spaced fish; (ii) strong background similarity (e.g., rocks, roots, or leaves with fish-like colors or textures); and (iii) distant targets in turbid water. Qwen2.5-VL exhibits basic zero-shot ability but degrades in these regimes—missing instances and placing keypoints off-body, which yields misaligned counts. Introducing our detect-to-count prompt (FishDetector-Base) enforces a box-first–then-keypoint policy, stabilizing localization and improving count consistency, though residual errors remain. After reinforcement fine-tuning, \textbf{FishDetector-R1} achieves substantial gains. Qualitatively, we observe fewer merges and splits in dense clusters, reduced background-induced false positives, and improved small-object recall—particularly for distant fish with very limited pixel extent. Keypoints also move off boundaries and onto the torso axis, increasing the fraction of points that lie within their matched boxes (or masks) and reducing keypoint–centroid dispersion; these effects align with lower absolute count error and higher match rate in our quantitative tables. The improved spatial alignment further stabilizes our weak-to-dense pipeline by providing reliable anchors for mask extraction in downstream segmentation. Remaining failure cases occur under extreme turbidity or near complete occlusion, where adjacent individuals can be under-separated; incorporating stronger separation priors or instance-aware refinement is a promising direction for future work.

\subsubsection{Zero-Shot Transfer}

\begin{figure*}[t]
    \centering
    \begin{subfigure}{1\linewidth}
        \centering
        \includegraphics[width=\linewidth]{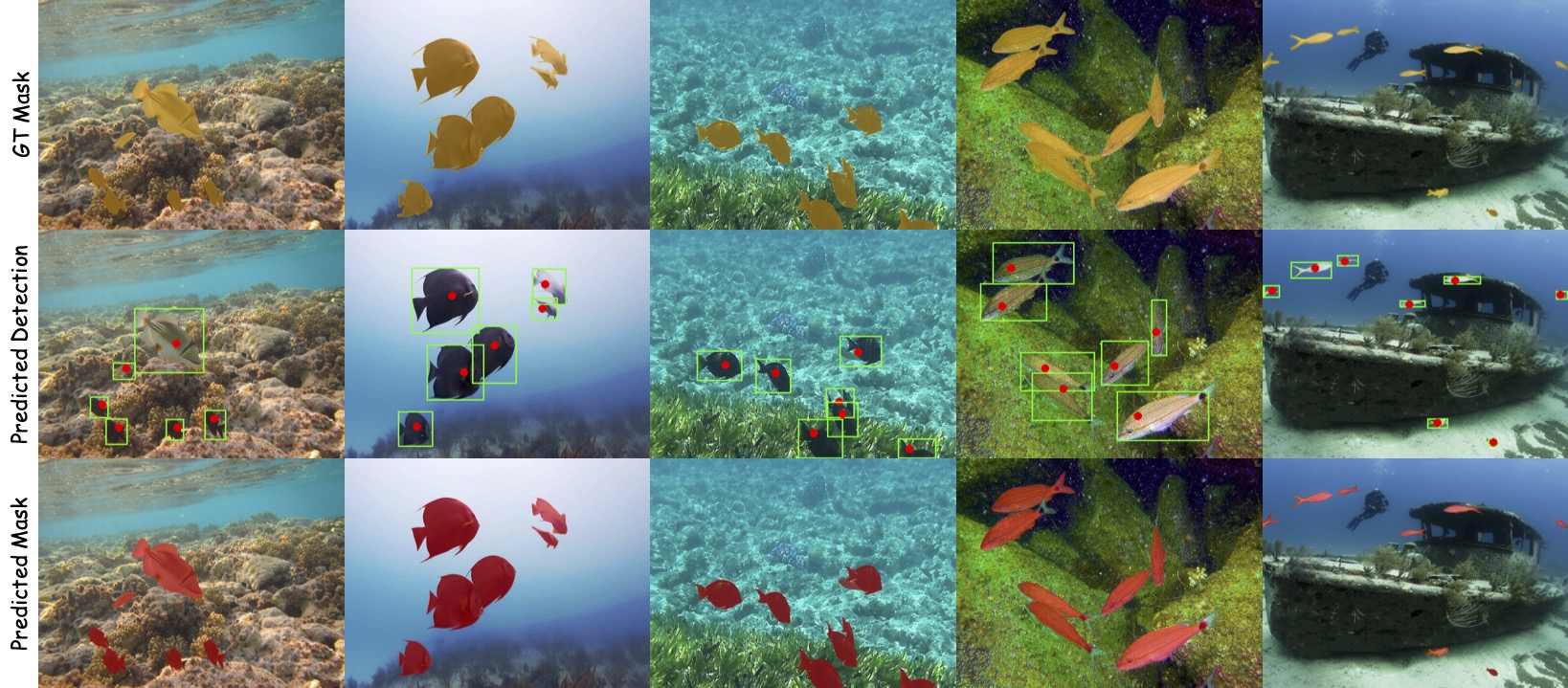}
        \caption{}
        \label{fig:top}
    \end{subfigure}

    \vspace{0.5em}

    \begin{subfigure}{1\linewidth}
        \centering
        \includegraphics[width=\linewidth]{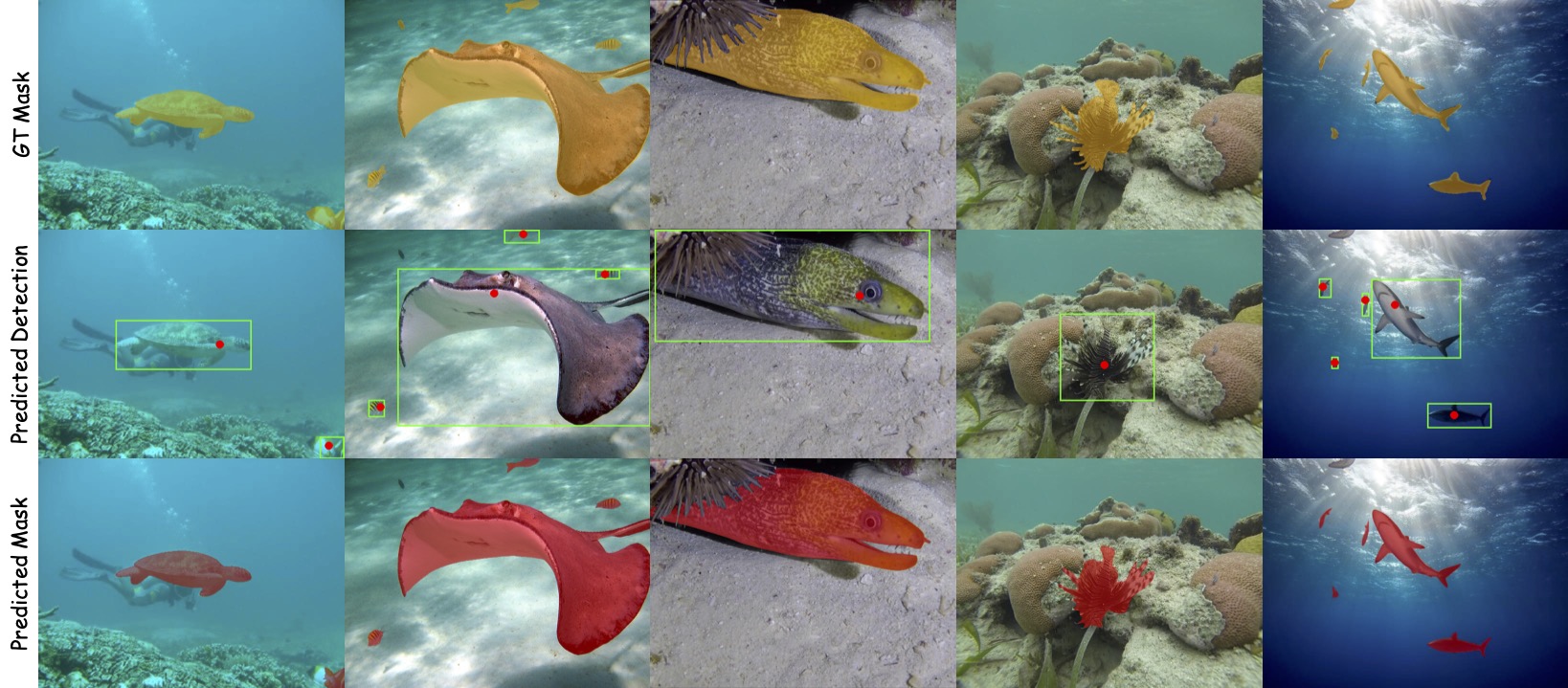}
        \caption{}
        \label{fig:bottom}
    \end{subfigure}

    \caption{
Qualitative results on the SUIM dataset~\cite{islam2020suim}. 
\textbf{(a)} Our model generalizes to the SUIM domain and successfully identifies fish species that share visual similarities with those in DeepFish. 
\textbf{(b)} Visualization of additional fish and vertebrate categories present in SUIM but absent in DeepFish, demonstrating the model's cross-domain adaptability and retention of original MLLM-level generalizability.
}
    \label{fig:suim_vertical_two}
\end{figure*}

In the main paper, we present quantitative results demonstrating that FishDetector\text{-}R1 generalizes effectively to SUIM~\cite{islam2020suim}, a dataset containing underwater scenes, lighting conditions, and fish species that differ substantially from those in DeepFish. To complement these findings, Figure~\ref{fig:suim_vertical_two} provides qualitative examples highlighting the model's zero-shot transfer capability. As shown in the figure, FishDetector\text{-}R1 successfully localizes and segments diverse fish species across a wide range of environments, including scenes with strong color cast, turbidity, occlusion, and cluttered coral backgrounds. Importantly, the model also handles non-fish marine animals (e.g., eels, rays) present in SUIM despite never encountering these categories during training, indicating strong cross-domain adaptability and retention of high-level semantic reasoning inherited from the underlying MLLM. The qualitative results further validate that FishDetector\text{-}R1 not only preserves its detection and segmentation fidelity in unseen domains but also maintains robustness to visual degradation and ecological variability, reinforcing its practical utility in real-world underwater monitoring scenarios.

\section{Potential Negative Societal Impact}
While FishDetector-R1 is intended for conservation and ecological monitoring, it carries non-trivial risks. Methodologically, domain bias, occlusion/turbidity, and reward shaping may yield over-confident counts or species misidentification; at scale, such errors could misinform stock assessments, quotas, or protected-area management. Application-wise, real-time localization of fish schools could be weaponized by industrial fleets or poachers to intensify harvest, exacerbate bycatch, or target threatened species, thereby harming ecosystems and small-scale fishers’ livelihoods. Large-scale camera deployment may disturb habitats (e.g., lighting or baiting) and enable surveillance of artisanal fishers or coastal communities, raising privacy and human-rights concerns; coastal video feeds could also be repurposed for broader monitoring beyond scientific aims. Finally, training and deployment incur energy costs with environmental externalities.

%% file: main.bbl
\begin{thebibliography}{41}
\providecommand{\natexlab}[1]{#1}
\providecommand{\url}[1]{\texttt{#1}}
\expandafter\ifx\csname urlstyle\endcsname\relax
  \providecommand{\doi}[1]{doi: #1}\else
  \providecommand{\doi}{doi: \begingroup \urlstyle{rm}\Url}\fi

\bibitem[Achiam et~al.(2023)Achiam, Adler, Agarwal, Ahmad, Akkaya, Aleman, Almeida, Altenschmidt, Altman, Anadkat, et~al.]{achiam2023gpt}
Josh Achiam, Steven Adler, Sandhini Agarwal, Lama Ahmad, Ilge Akkaya, Florencia~Leoni Aleman, Diogo Almeida, Janko Altenschmidt, Sam Altman, Shyamal Anadkat, et~al.
\newblock Gpt-4 technical report.
\newblock \emph{arXiv preprint arXiv:2303.08774}, 2023.

\bibitem[Al~Muksit et~al.(2022)Al~Muksit, Hasan, Emon, Haque, Anwary, and Shatabda]{al2022yolo}
Abdullah Al~Muksit, Fakhrul Hasan, Md~Fahad Hasan~Bhuiyan Emon, Md~Rakibul Haque, Arif~Reza Anwary, and Swakkhar Shatabda.
\newblock Yolo-fish: A robust fish detection model to detect fish in realistic underwater environment.
\newblock \emph{Ecological Informatics}, 72:\penalty0 101847, 2022.

\bibitem[Bai et~al.(2025)Bai, Chen, Liu, Wang, Ge, Song, Dang, Wang, Wang, Tang, et~al.]{bai2025qwen2}
Shuai Bai, Keqin Chen, Xuejing Liu, Jialin Wang, Wenbin Ge, Sibo Song, Kai Dang, Peng Wang, Shijie Wang, Jun Tang, et~al.
\newblock Qwen2. 5-vl technical report.
\newblock \emph{arXiv preprint arXiv:2502.13923}, 2025.

\bibitem[Bearman et~al.(2016)Bearman, Russakovsky, Ferrari, and Fei-Fei]{bearman2016s}
Amy Bearman, Olga Russakovsky, Vittorio Ferrari, and Li Fei-Fei.
\newblock What’s the point: Semantic segmentation with point supervision.
\newblock In \emph{European conference on computer vision}, pages 549--565. Springer, 2016.

\bibitem[Cai and Vasconcelos(2018)]{cai2018cascade}
Zhaowei Cai and Nuno Vasconcelos.
\newblock Cascade r-cnn: Delving into high quality object detection.
\newblock In \emph{Proceedings of the IEEE conference on computer vision and pattern recognition}, pages 6154--6162, 2018.

\bibitem[Comanici et~al.(2025)Comanici, Bieber, Schaekermann, Pasupat, Sachdeva, Dhillon, Blistein, Ram, Zhang, Rosen, et~al.]{comanici2025gemini}
Gheorghe Comanici, Eric Bieber, Mike Schaekermann, Ice Pasupat, Noveen Sachdeva, Inderjit Dhillon, Marcel Blistein, Ori Ram, Dan Zhang, Evan Rosen, et~al.
\newblock Gemini 2.5: Pushing the frontier with advanced reasoning, multimodality, long context, and next generation agentic capabilities.
\newblock \emph{arXiv preprint arXiv:2507.06261}, 2025.

\bibitem[Garcia et~al.(2020)Garcia, Prados, Quintana, Tempelaar, Gracias, Rosen, V{\aa}gst{\o}l, and L{\o}vall]{garcia2020automatic}
Rafael Garcia, Ricard Prados, Josep Quintana, Alexander Tempelaar, Nuno Gracias, Shale Rosen, H{\aa}vard V{\aa}gst{\o}l, and Kristoffer L{\o}vall.
\newblock Automatic segmentation of fish using deep learning with application to fish size measurement.
\newblock \emph{ICES Journal of Marine Science}, 77\penalty0 (4):\penalty0 1354--1366, 2020.

\bibitem[Guerrero-G{\'o}mez-Olmedo et~al.(2015)Guerrero-G{\'o}mez-Olmedo, Torre-Jim{\'e}nez, L{\'o}pez-Sastre, Maldonado-Basc{\'o}n, and Onoro-Rubio]{guerrero2015extremely}
Ricardo Guerrero-G{\'o}mez-Olmedo, Beatriz Torre-Jim{\'e}nez, Roberto L{\'o}pez-Sastre, Saturnino Maldonado-Basc{\'o}n, and Daniel Onoro-Rubio.
\newblock Extremely overlapping vehicle counting.
\newblock In \emph{Iberian conference on pattern recognition and image analysis}, pages 423--431. Springer, 2015.

\bibitem[Guo et~al.(2025)Guo, Yang, Zhang, Song, Zhang, Xu, Zhu, Ma, Wang, Bi, et~al.]{guo2025deepseek}
Daya Guo, Dejian Yang, Haowei Zhang, Junxiao Song, Ruoyu Zhang, Runxin Xu, Qihao Zhu, Shirong Ma, Peiyi Wang, Xiao Bi, et~al.
\newblock Deepseek-r1: Incentivizing reasoning capability in llms via reinforcement learning.
\newblock \emph{arXiv preprint arXiv:2501.12948}, 2025.

\bibitem[He et~al.(2016)He, Zhang, Ren, and Sun]{he2016deep}
Kaiming He, Xiangyu Zhang, Shaoqing Ren, and Jian Sun.
\newblock Deep residual learning for image recognition.
\newblock In \emph{Proceedings of the IEEE conference on computer vision and pattern recognition}, pages 770--778, 2016.

\bibitem[Hong et~al.(2024)Hong, Zhou, Hua, Lv, and Dong]{hong2024watersam}
Yang Hong, Xiaowei Zhou, Ruzhuang Hua, Qingxuan Lv, and Junyu Dong.
\newblock Watersam: Adapting sam for underwater object segmentation.
\newblock \emph{Journal of Marine Science and Engineering}, 12\penalty0 (9):\penalty0 1616, 2024.

\bibitem[Hurst et~al.(2024)Hurst, Lerer, Goucher, Perelman, Ramesh, Clark, Ostrow, Welihinda, Hayes, Radford, et~al.]{hurst2024gpt}
Aaron Hurst, Adam Lerer, Adam~P Goucher, Adam Perelman, Aditya Ramesh, Aidan Clark, AJ Ostrow, Akila Welihinda, Alan Hayes, Alec Radford, et~al.
\newblock Gpt-4o system card.
\newblock \emph{arXiv preprint arXiv:2410.21276}, 2024.

\bibitem[Islam et~al.(2020)Islam, Edge, Xiao, Luo, Mehtaz, Morse, Enan, and Sattar]{islam2020suim}
Md~Jahidul Islam, Chelsey Edge, Yuyang Xiao, Peigen Luo, Muntaqim Mehtaz, Christopher Morse, Sadman~Sakib Enan, and Junaed Sattar.
\newblock Semantic segmentation of underwater imagery: Dataset and benchmark.
\newblock In \emph{2020 IEEE/RSJ international conference on intelligent robots and systems (IROS)}, pages 1769--1776. IEEE, 2020.

\bibitem[Kirillov et~al.(2023)Kirillov, Mintun, Ravi, Mao, Rolland, Gustafson, Xiao, Whitehead, Berg, Lo, et~al.]{kirillov2023segment}
Alexander Kirillov, Eric Mintun, Nikhila Ravi, Hanzi Mao, Chloe Rolland, Laura Gustafson, Tete Xiao, Spencer Whitehead, Alexander~C Berg, Wan-Yen Lo, et~al.
\newblock Segment anything.
\newblock In \emph{ICCV}, pages 4015--4026, 2023.

\bibitem[Laradji et~al.(2020)Laradji, Saleh, Rodriguez, Nowrouzezahrai, Azghadi, and Vazquez]{laradji2020affinity}
Issam Laradji, Alzayat Saleh, Pau Rodriguez, Derek Nowrouzezahrai, Mostafa~Rahimi Azghadi, and David Vazquez.
\newblock Affinity lcfcn: Learning to segment fish with weak supervision.
\newblock \emph{arXiv preprint arXiv:2011.03149}, 2020.

\bibitem[Laradji et~al.(2018)Laradji, Rostamzadeh, Pinheiro, Vazquez, and Schmidt]{laradji2018blobs}
Issam~H Laradji, Negar Rostamzadeh, Pedro~O Pinheiro, David Vazquez, and Mark Schmidt.
\newblock Where are the blobs: Counting by localization with point supervision.
\newblock In \emph{Proceedings of the european conference on computer vision (ECCV)}, pages 547--562, 2018.

\bibitem[Lian et~al.(2023)Lian, Li, Cong, Li, Zhang, and Kwong]{lian2023watermask}
Shijie Lian, Hua Li, Runmin Cong, Suqi Li, Wei Zhang, and Sam Kwong.
\newblock Watermask: Instance segmentation for underwater imagery.
\newblock In \emph{ICCV}, pages 1305--1315, 2023.

\bibitem[Lin et~al.(2014)Lin, Maire, Belongie, Hays, Perona, Ramanan, Doll{\'a}r, and Zitnick]{lin2014microsoft}
Tsung-Yi Lin, Michael Maire, Serge Belongie, James Hays, Pietro Perona, Deva Ramanan, Piotr Doll{\'a}r, and C~Lawrence Zitnick.
\newblock Microsoft coco: Common objects in context.
\newblock In \emph{European conference on computer vision}, pages 740--755. Springer, 2014.

\bibitem[Liu et~al.(2024)Liu, Zeng, Ren, Li, Zhang, Yang, Jiang, Li, Yang, Su, et~al.]{liu2024grounding}
Shilong Liu, Zhaoyang Zeng, Tianhe Ren, Feng Li, Hao Zhang, Jie Yang, Qing Jiang, Chunyuan Li, Jianwei Yang, Hang Su, et~al.
\newblock Grounding dino: Marrying dino with grounded pre-training for open-set object detection.
\newblock In \emph{European conference on computer vision}, pages 38--55. Springer, 2024.

\bibitem[Liu et~al.(2025{\natexlab{a}})Liu, Peng, Zhong, Yue, Lu, Yu, and Jia]{seg-zero}
Yuqi Liu, Bohao Peng, Zhisheng Zhong, Zihao Yue, Fanbin Lu, Bei Yu, and Jiaya Jia.
\newblock Seg-zero: Reasoning-chain guided segmentation via cognitive reinforcement.
\newblock \emph{arXiv preprint arXiv:2503.06520}, 2025{\natexlab{a}}.

\bibitem[Liu et~al.(2025{\natexlab{b}})Liu, Qu, Zhong, Peng, Liu, Yu, and Jia]{visionreasoner}
Yuqi Liu, Tianyuan Qu, Zhisheng Zhong, Bohao Peng, Shu Liu, Bei Yu, and Jiaya Jia.
\newblock Visionreasoner: Unified visual perception and reasoning via reinforcement learning.
\newblock \emph{arXiv preprint arXiv:2505.12081}, 2025{\natexlab{b}}.

\bibitem[Liu et~al.(2025{\natexlab{c}})Liu, Sun, Zang, Dong, Cao, Duan, Lin, and Wang]{visual-rft}
Ziyu Liu, Zeyi Sun, Yuhang Zang, Xiaoyi Dong, Yuhang Cao, Haodong Duan, Dahua Lin, and Jiaqi Wang.
\newblock Visual-rft: Visual reinforcement fine-tuning.
\newblock \emph{arXiv preprint arXiv:2503.01785}, 2025{\natexlab{c}}.

\bibitem[Medeiros(2025)]{medeiros2025langsegmentanything}
Luca Medeiros.
\newblock lang-segment-anything: Sam with text prompt.
\newblock \url{https://github.com/luca-medeiros/lang-segment-anything}, 2025.
\newblock GitHub repository, accessed 2025-11.

\bibitem[OpenAI(2025)]{openai_gpt41_2025}
OpenAI.
\newblock Introducing {GPT-4.1} in the api.
\newblock \url{https://openai.com/index/gpt-4-1/}, 2025.
\newblock Accessed: 2025-09-16.

\bibitem[Qin et~al.(2016)Qin, Li, Liang, Peng, and Zhang]{qin2016deepfish}
Hongwei Qin, Xiu Li, Jian Liang, Yigang Peng, and Changshui Zhang.
\newblock Deepfish: Accurate underwater live fish recognition with a deep architecture.
\newblock \emph{Neurocomputing}, 187:\penalty0 49--58, 2016.

\bibitem[Rafailov et~al.(2023)Rafailov, Sharma, Mitchell, Manning, Ermon, and Finn]{DPO}
Rafael Rafailov, Archit Sharma, Eric Mitchell, Christopher~D Manning, Stefano Ermon, and Chelsea Finn.
\newblock Direct preference optimization: Your language model is secretly a reward model.
\newblock \emph{Advances in neural information processing systems}, 36:\penalty0 53728--53741, 2023.

\bibitem[Ravi et~al.(2024)Ravi, Gabeur, Hu, Hu, Ryali, Ma, Khedr, R{\"a}dle, Rolland, Gustafson, et~al.]{ravi2024sam}
Nikhila Ravi, Valentin Gabeur, Yuan-Ting Hu, Ronghang Hu, Chaitanya Ryali, Tengyu Ma, Haitham Khedr, Roman R{\"a}dle, Chloe Rolland, Laura Gustafson, et~al.
\newblock Sam 2: Segment anything in images and videos.
\newblock \emph{arXiv preprint arXiv:2408.00714}, 2024.

\bibitem[Saleh et~al.(2020)Saleh, Laradji, Konovalov, Bradley, Vazquez, and Sheaves]{saleh2020realistic}
Alzayat Saleh, Issam~H Laradji, Dmitry~A Konovalov, Michael Bradley, David Vazquez, and Marcus Sheaves.
\newblock A realistic fish-habitat dataset to evaluate algorithms for underwater visual analysis.
\newblock \emph{Scientific reports}, 10\penalty0 (1):\penalty0 14671, 2020.

\bibitem[Schulman et~al.(2017)Schulman, Wolski, Dhariwal, Radford, and Klimov]{PPO}
John Schulman, Filip Wolski, Prafulla Dhariwal, Alec Radford, and Oleg Klimov.
\newblock Proximal policy optimization algorithms.
\newblock \emph{arXiv preprint arXiv:1707.06347}, 2017.

\bibitem[Shao et~al.(2022)Shao, Chen, Shao, Ji, Xiao, Ye, Zhuang, and Xiao]{shao2022deep}
Feifei Shao, Long Chen, Jian Shao, Wei Ji, Shaoning Xiao, Lu Ye, Yueting Zhuang, and Jun Xiao.
\newblock Deep learning for weakly-supervised object detection and localization: A survey.
\newblock \emph{Neurocomputing}, 496:\penalty0 192--207, 2022.

\bibitem[Song et~al.(2024)Song, Bagoren, Andigani, Sethuraman, and Skinner]{song2024turtlmap}
Jingyu Song, Onur Bagoren, Razan Andigani, Advaith Sethuraman, and Katherine~A Skinner.
\newblock Turtlmap: Real-time localization and dense mapping of low-texture underwater environments with a low-cost unmanned underwater vehicle.
\newblock In \emph{2024 IEEE/RSJ International Conference on Intelligent Robots and Systems (IROS)}, pages 1191--1198. IEEE, 2024.

\bibitem[Song et~al.(2025)Song, Ma, Bagoren, Sethuraman, Zhang, and Skinner]{song2025oceansim}
Jingyu Song, Haoyu Ma, Onur Bagoren, Advaith~V. Sethuraman, Yiting Zhang, and Katherine~A. Skinner.
\newblock Oceansim: A gpu-accelerated underwater robot perception simulation framework.
\newblock In \emph{2025 IEEE/RSJ International Conference on Intelligent Robots and Systems (IROS)}. IEEE, 2025.

\bibitem[Stankus(2021)]{stankus2021state}
Austin Stankus.
\newblock State of world aquaculture 2020 and regional reviews: Fao webinar series.
\newblock \emph{FAO aquaculture newsletter}, \penalty0 (63):\penalty0 17--18, 2021.

\bibitem[Team et~al.(2023)Team, Anil, Borgeaud, Alayrac, Yu, Soricut, Schalkwyk, Dai, Hauth, Millican, et~al.]{team2023gemini}
Gemini Team, Rohan Anil, Sebastian Borgeaud, Jean-Baptiste Alayrac, Jiahui Yu, Radu Soricut, Johan Schalkwyk, Andrew~M Dai, Anja Hauth, Katie Millican, et~al.
\newblock Gemini: a family of highly capable multimodal models.
\newblock \emph{arXiv preprint arXiv:2312.11805}, 2023.

\bibitem[Touvron et~al.(2023)Touvron, Lavril, Izacard, Martinet, Lachaux, Lacroix, Rozi{\`e}re, Goyal, Hambro, Azhar, et~al.]{touvron2023llama}
Hugo Touvron, Thibaut Lavril, Gautier Izacard, Xavier Martinet, Marie-Anne Lachaux, Timoth{\'e}e Lacroix, Baptiste Rozi{\`e}re, Naman Goyal, Eric Hambro, Faisal Azhar, et~al.
\newblock Llama: Open and efficient foundation language models.
\newblock \emph{arXiv preprint arXiv:2302.13971}, 2023.

\bibitem[Xu et~al.(2023)Xu, Su, and Liu]{xu2023aquasam}
Muduo Xu, Jianhao Su, and Yutao Liu.
\newblock Aquasam: Underwater image foreground segmentation.
\newblock In \emph{International Forum on Digital TV and Wireless Multimedia Communications}, pages 3--14, 2023.

\bibitem[You and Wu(2025)]{seg-r1}
Zuyao You and Zuxuan Wu.
\newblock Seg-r1: Segmentation can be surprisingly simple with reinforcement learning.
\newblock \emph{arXiv preprint arXiv:2506.22624}, 2025.

\bibitem[Yu et~al.(2025)Yu, Lin, Zhao, Yin, Wei, Peng, Wei, Sun, Han, Ge, et~al.]{yu2025perception}
En Yu, Kangheng Lin, Liang Zhao, Jisheng Yin, Yana Wei, Yuang Peng, Haoran Wei, Jianjian Sun, Chunrui Han, Zheng Ge, et~al.
\newblock Perception-r1: Pioneering perception policy with reinforcement learning.
\newblock \emph{arXiv preprint arXiv:2504.07954}, 2025.

\bibitem[Zhang et~al.(2022)Zhang, Wu, and Bao]{zhang2022dpanet}
Wenbo Zhang, Chaoyi Wu, and Zhenshan Bao.
\newblock Dpanet: dual pooling-aggregated attention network for fish segmentation.
\newblock \emph{IET computer vision}, 16\penalty0 (1):\penalty0 67--82, 2022.

\bibitem[Zhang et~al.(2025)Zhang, Dong, Zhang, Zhao, Zhou, Xi, Jin, Fan, Zhou, Fu, et~al.]{zhang2025reinforcement}
Zhihao Zhang, Qiaole Dong, Qi Zhang, Jun Zhao, Enyu Zhou, Zhiheng Xi, Senjie Jin, Xiaoran Fan, Yuhao Zhou, Yanwei Fu, et~al.
\newblock Reinforcement fine-tuning enables mllms learning novel tasks stably.
\newblock \emph{arXiv preprint arXiv:2506.23508}, 2025.

\bibitem[Zheng et~al.(2025)Zheng, Lu, Wang, Feng, Kuang, and Xiong]{zheng2025easyr1}
Yaowei Zheng, Junting Lu, Shenzhi Wang, Zhangchi Feng, Dongdong Kuang, and Yuwen Xiong.
\newblock Easyr1: An efficient, scalable, multi-modality rl training framework.
\newblock \url{https://github.com/hiyouga/EasyR1}, 2025.

\end{thebibliography}
